\documentclass[twoside]{article}
\usepackage{subfigure,graphicx}
\usepackage{amsmath,amsfonts,amssymb}
\usepackage{rotating}
\usepackage[hyperfootnotes=false]{hyperref}
\usepackage{amsthm}
\usepackage[ruled,vlined]{algorithm2e}
\usepackage[table]{xcolor}
\usepackage{color,colortbl,tabularx}
\usepackage[english]{babel}
\usepackage{booktabs}
\usepackage{algpseudocode} % For algorithmicx
\usepackage{float}
\usepackage{xcolor}
\usepackage{subcaption}
\usepackage{multirow}
\usepackage[accepted]{aistats2025}
\definecolor{darkgreen}{rgb}{0.0, 0.2, 0.0}

\hypersetup{colorlinks,linkcolor={black},citecolor={darkgreen}, urlcolor={darkgreen}}
\usepackage[round]{natbib}

\begin{document}


% \runningauthor{Ange-Clément Akazan, Ioannis Mitliagkas, Alexia Jolicoeur-Martineau}

\twocolumn[
\aistatstitle{Generating Tabular Data Using Heterogeneous Sequential Feature Forest Flow Matching}
%     \begin{tabular}{p{0.3\textwidth} p{0.3\textwidth} p{0.3\textwidth}}
%         \small $^1$Ange-Clément Akazan  & \small Ioannis Mitliagkas & \small Alexia Jolicoeur-Martineau\\
%         \small AIMS Research and Innovation Center, Kigali, Rwanda&  \small $^1$ Mila Québec AI Institute, Université de Montréal, Canada    &\small SAIT Montreal AI Lab, Montréal, Canada \\ \\
%     \end{tabular}

    % \\\footnote{This work was partially conducted during an internship at the Montreal Institute for Learning Algorithms (MILA).}
    %  \begin{tabular}{p{0.3\textwidth} p{0.3\textwidth} p{0.3\textwidth}}
    %   \\
    % \end{tabular}
\aistatsauthor{\small Ange-Clément Akazan$^{1}$ \And Ioannis Mitliagkas \And Alexia Jolicoeur-Martineau }
 \aistatsaddress{  \small AIMS Research and \\Innovation Center, Kigali, Rwanda \And  $^1$Mila Québec AI Institute, \\Université de Montréal, Canada \And     \small SAIT Montreal AI Lab, \\Montréal, Canada }
 % \footnote{This work was partially conducted during an internship at the Montreal Institute for Learning Algorithms (MILA).}
% \vspace{2em}
]
\begin{abstract}
%General Introduction
Privacy and regulatory constraints make data generation vital to advancing machine learning without relying on real-world datasets. A leading approach for tabular data generation is the Forest Flow (FF) method, which combines Flow Matching with XGBoost. 
%Limitation of the existing method
Despite its good performance, FF is slow and makes errors when treating categorical variables as one-hot continuous features. It is also highly sensitive to small changes in the initial conditions of the ordinary differential equation (ODE). To overcome these limitations, we develop  Heterogeneous Sequential Feature Forest Flow (HS3F).
 %Methods
Our method generates data sequentially (feature-by-feature), reducing the dependency on noisy initial conditions through the additional information from previously generated features.
Furthermore, it generates categorical variables using multinomial sampling (from an XGBoost classifier) instead of flow matching, improving generation speed.  We also use a Runge-Kutta 4th order (Rg4) ODE solver for improved performance over the Euler solver used in FF. 
%Results 
Our experiments with 25 datasets reveal that HS3F produces higher quality and more diverse synthetic data than FF, especially for categorical variables. It also generates data 21-27 times faster for datasets with $\geq 20\%$ categorical variables. HS3F further demonstrates enhanced robustness to affine transformation in flow ODE initial conditions compared to FF.
%Conclusion and recommendations
This study not only validates the HS3F but also unveils promising new strategies to advance generative models.
\end{abstract}

\section{\bf Introduction}
Tabular datasets, typically derived from surveys and experiments, are crucial in various fields. They support decision-making in finance, healthcare, climate science, economics, and social sciences by providing detailed and quantifiable insights necessary for informed analysis and strategy \citep{9998482}. Generating tabular data is an important task as it improves the performance of machine learning models through data augmentation \citep{MOTAMED2021100779}, mitigates bias by addressing class imbalance \citep{JUWARA2024100946}, and enhances data security \citep{NEURIPS2021_22456f4b}, enabling institutions to share valuable information without compromising privacy.

However, as discussed in several papers, including \citep{kim2023stasy,10.5555/3618408.3619133,jolicoeurmartineau2024generating}, the generation of tabular data presents unique challenges that are often more complex than those associated with other data types. %These challenges contribute to the relative scarcity %of studies focused on this problem. 
Key issues related to tabular data include feature heterogeneity (the coexistence of continuous and categorical features), the typically small size of tabular datasets, the low signal-to-noise ratio, and missing values.
\begin{figure*}[ht]
    \centering
    \includegraphics[width=0.7\linewidth]{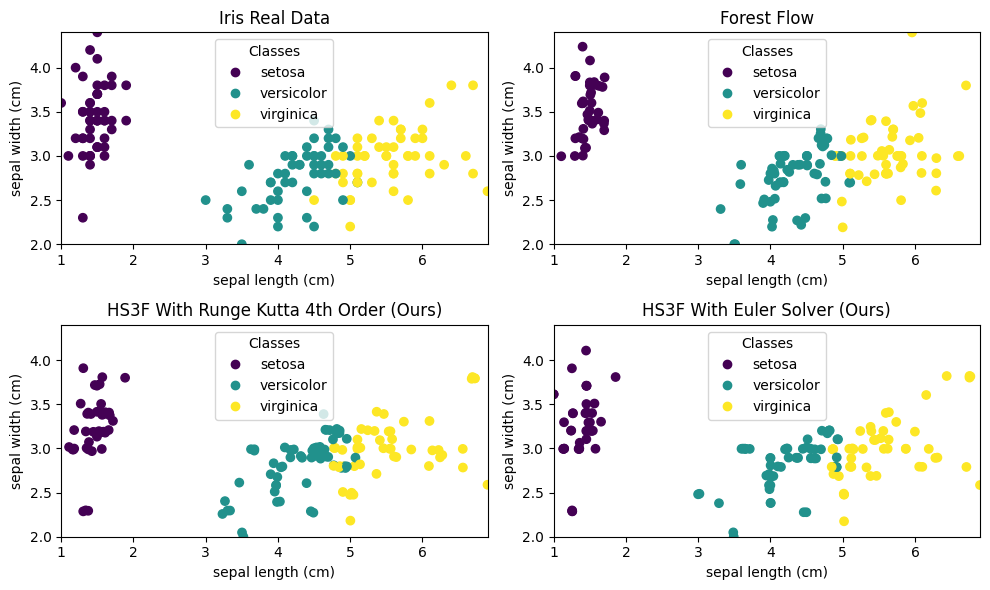}
    \caption{Iris Data: Three-Way Interaction Plot of Sepal Width vs. Length by Species, Comparing Real Data with Data Generated by ForestFlow, \citep{jolicoeurmartineau2024generating}, and by H3SF  Using Euler and Runge Kutta 4th Order Solvers}
    \label{Iris_plot}
\end{figure*}

Interestingly, significant advancements in AI have been made, especially in synthetic data generation, with the deployment of deep-generative models (DGMs), which are deep learning models capable of learning the underlying distribution and relationships within data, enabling the generation of new data mirroring these learned patterns. 
%DGMs were used to tackle tabular data generation 
%continuous normalizing flow \cite{rubanova2019latent},
DGMs such as generative adversarial networks (GANs) \citep{goodfellow2014generative}, variational autoencoders (VAEs)\citep{kingma2013auto},  diffusion models \citep{pmlr-v37-sohl-dickstein15,NEURIPS2020_4c5bcfec,song2021scorebasedgenerativemodelingstochastic} and conditional flow matching (CFM)\citep{lipman2023flow,tong2023improving}, were  successfully used to handle tabular data generation.

One of the recent compelling methods for tabular data generation is conditional flow matching (CFM). CFM is an emerging generative modeling framework that is gaining prominence in AI-driven data generation, whose goal is to determine a flow that pushes a noisy sample distribution into the original data distribution. This process involves estimating vector fields conditioned on a data sample using a neural network, which, via an ODE, guides the trajectory of a flow that pushes the noise distribution to the target distribution. 
% Once these vector fields are accurately estimated, they are used to solve an  that defines the flow, effectively transforming the noisy data distribution back to the original data distribution. 

Motivated by CFM, \cite{jolicoeurmartineau2024generating}  deployed ForestFlow Matching (FF), a CFM-based tabular data generation method that uses eXtreme Gradient Boosting regressor models (XGBoost regressor \citep{chen2016xgboost}) for estimating conditional velocity vector fields instead of the typical neural networks. This non-deep-learning method produces highly realistic synthetic tabular data, even when the training dataset contains missing values. In an extensive study, the authors showed that FF outperforms many of the most effective and widely used deep-learning generative models while running with limited hardware (around 12-24 CPU cores, as found in a modern laptop).

Despite its efficiency,  FF  is slow and  very sensitive to changes in the flow ODE initial condition (when the initial condition data distribution slightly differs from the Standard Gaussian). Additionally, FF  does not effectively address feature heterogeneity. It handles discrete data by doing a one-hot encoding of the categories in order to relax the categorical distribution space into a continuous space. This solution is motivated by the fact that conditional flow matching is designed for typical continuous spaces.
However, this solution leads to approximation errors for categorical data. As a result, FF tends to perform less effectively when there is a large number of categorical features in the data set.  

Our method, heterogeneous sequential feature forest Flow (HS3F), extends ForestFlow \citep{jolicoeurmartineau2024generating} through an explicit mechanism to handle heterogeneity. HS3F processes data sequentially (feature after feature). This allows us to generate continuous features using FF, but handle categorical features using a (Xgboost) classifier. Categorical features are generated by randomly sampling from the probabilities of the learned classifier, which massively improves speed and closeness to the real data distribution.

Our main contributions can be listed as follows:
\begin{itemize}
 \item A novel and efficient Forest Flow framework denoted as heterogeneous sequential feature forest Flow (HS3F) that generates data one feature after the other, leveraging information from previously generated features. 

\item A natural, multinomial sampling for categorical features, based on the Xgboost classifier's predicted class probability. 
This improvement avoids lifting the categorical space into a continuous, higher-dimensional space.
Continuous features are generated 
via flow matching.

\item An extensive data generation study (across 25 datasets) using various metrics. We demonstrate that HS3F produces higher quality and more diverse
synthetic data than FF while significantly reducing computational time, especially when the data contains categorical features.  

\item An experimental demonstration that HS3F is more robust to distributional change in the flow ODE's initial condition. 
\end{itemize}
\begin{figure*}[ht!]
    \centering
\includegraphics[width=1.0\linewidth]{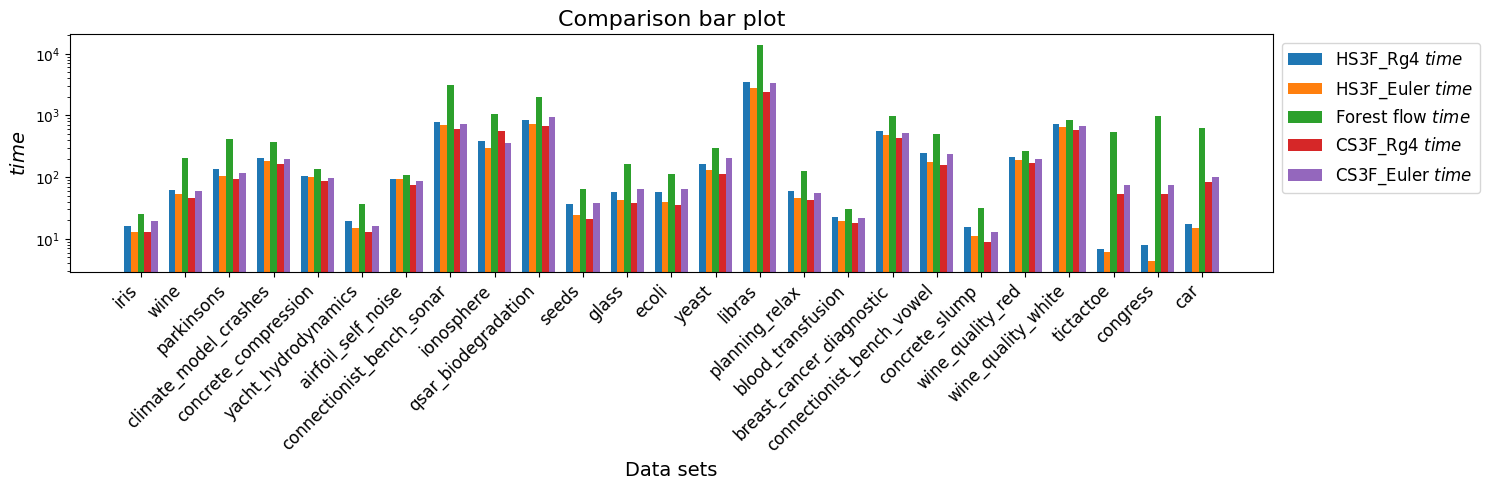}
    \caption{Data generation time comparison per models across generated datasets. We see that 
    our proposed method HS3F is much faster than CS3F and Forest Flow, especially for datasets with many categorical features.}
    \label{fig6_1}
\end{figure*}
\subsection{Related Work}
Deep generative models have revolutionized tabular data generation by learning complex, nonlinear representations to generate highly realistic synthetic data. Among these, Generative Adversarial Networks \citep{goodfellow2014generative} (GANs) have been adapted for tabular data generation, leveraging architectures like Deep GANs, conditional GANs, and modified GANs to enhance data quality, privacy, and distribution preservation \citep{ Park_2018, esteban2017realvalued, NEURIPS2019__254ed7d2, kim2021octgan, pmlr-v157-zhao21a, yoon2018pategan}. Variational Autoencoders \citep{kingma2013auto} (VAEs) have also made significant contributions through extensions like TVAE, which employs a triple-loss objective to improve generation quality \citep{ ishfaq2023tvae}. 
% \newline \newline
In recent advancements, diffusion-based models \citep{pmlr-v37-sohl-dickstein15, song2021scorebasedgenerativemodelingstochastic} and flow matching models \citep{lipman2023flow} have emerged as powerful tools for tabular data generation, often surpassing GANs and VAEs in performance. For example, the $TABDDM$ model utilizes Gaussian and multinomial diffusion for effective continuous and categorical data generation \citep{10.5555/3618408.3619133}. The score-based diffusion model framework was also successfully adapted for tabular data generation through  $STaSy$, which additionally employs self-paced learning to gain efficiency and better performance than $TABDDM$ \citep{song2021scorebasedgenerativemodelingstochastic, kim2023stasy}.
Conditional Flow Matching (CFM) \citep{lipman2023flow} represents a newer approach that offers competitive performance and speed. In particular, \cite{jolicoeurmartineau2024generating} introduced the first flow-based model for tabular data generation using XGBoost regressors, known as Forest Flow (FF). Forest Flow (FF) has established itself as one of the best method in tabular data generation, showing exceptional performance in extensive studies across a wide range of datasets.  Inspired by  FF, we propose novel methods to advance this idea further.

\paragraph{}The rest of this study is organized as follows.  Section 2 presents the background of our study, including an introduction to flow matching and its variants. In Section 3, we discuss the methods used in this study, elaborating on the motivations behind their selection, and explaining each method. Section 4 presents the results obtained from our experiments. Finally, Section 5 concludes the study and suggests potential directions for future work that could make valuable contributions. Additional details and supplementary material are provided in the Appendix.
\section{Background}\label{background}
This section introduces the flow matching framework, provides details on conditional flow matching and independent coupling flow matching, and discusses FF, the method that inspired our approach.

\subsection{Conditional Flow Matching (CFM) Framework}
\subsubsection{Flow Matching}
Let us assume that the data space is $\mathbb{R}^n$, with data points $x_1$ following the unknown distribution $q_1$,  and the noisy data point $x_0 $ following the distribution $p_0$. The motivation behind flow matching is to find a time-dependent probability path $p_t, \, t\in[0,1]$ such that $p_{t=0}=p_0$ and $p_{t=1}=q_1$, assuming that this probability path is defined  by a time-dependent map $\phi_t$ called flow which is induced by a time-dependent vector $v_t$ called velocity vector field, through the following  flow ODE system (see Eq (\ref{vct}))\begin{equation} \dfrac{\partial \phi_t (x)}{dt}= v_t(\phi_t(x)); \,\,  \phi_0(x_0)=x_0\sim p_0, \, t\in[0,1].\label{vct}\end{equation}  $\phi_t$ determines the path $p_t$
via the relationship  $p_{t}=[\phi_t]_{\#}p_0$ where $[]_{\#}$ is the push-forward relationship.
% (see Appendix Eq.\ref{flw} )
The core concept of flow matching is the velocity vector field $v_t$, as it uniquely determines $\phi_t$ and therefore $p_t$ so that we can sample realistic data $\hat{x}_1 \sim p_1$ given any $z \sim p_0$.
Given a velocity vector $v_t$, the main objective of flow matching is then to regress a neural network  \( v^{\theta}_t: [0, 1] \times \mathbb{R}^d \rightarrow \mathbb{R}^d
\) against the velocity vector field $v_t$ by minimizing the loss function $\mathcal{L}_{FM}$  defined as follows:
 \begin{equation}
    \mathcal{L}_{FM}(\theta) = E_{t, p_t(x)} \|v_t(x_t) - \hat{v}^\theta_t(x)\|^2 
\label{flwObj}.
\end{equation}
After training, $\hat{v}^\theta_t$ is used  to numerically solve  the ODE (Eq.eq\ref{vct}) in order to determine the  flow estimate $\hat{\phi}_t$ that pushes, through the push-forward relationship(Eq.\eqref{flw}), $p_0$ to   $q_1$. 

Unfortunately, this flow-matching training objective is intractable because we do not have knowledge of suitable $v_t$ and $p_t$. \footnote{More details about the conditional flow matching framework can be found in \cite{lipman2023flow, tong2023improving}}

\paragraph{Conditional Flow Matching}To bypass this aforementioned tractability challenge,  \cite{lipman2023flow} suggested to instead minimize the conditional flow matching objective $\mathcal{L}_{CFM}$ (see \ref{CondflwObj}) and demonstrated that $\mathcal{L}_{CFM}$ and  $\mathcal{L}_{FM}$(Eq.\eqref{flwObj}) have the same gradient:
\begin{equation}
    \mathcal{L}_{CFM}(\theta) = E_{t, q(x_1), p_t(x|x_1)} \|v^{\theta}_t(\phi_t(x|x_1))-v_t(x|x_1)\|^2,
\label{CondflwObj}
\end{equation}
where $q$ is some uniform distribution over the original data $x_1$, $p_t(x|x_1)$ is a distribution equal to $p_0$ at time $t=0$ and centered in $x_1$ at time $t=1$ with mean $\mu_t$ and standard deviation $\sigma_t$,  $v_t(x|x_1)$ is the conditional velocity vector that induced $p_t(x|x_1)$ via the linear conditional flow $x_t=\phi_t(x|x_1)=\mu_t(x_1)+x\sigma_t(x_1)$.
 %\ref{liu2022rectified} 

\paragraph{Independent Coupling Flow Matching:}\cite{tong2023improving} proposed the independent coupling flow matching (ICFM), a modified conditional flow matching method that provides better and faster flow directions. The ICFM framework assumes that  $x_0$  and $x_1$ are sampled independently, $p_t(x|x_0,x_1)$ is a  Gaussian distribution with expectation $\mu_t(x_0,x_1)=(1-t)x_0+tx_1$ and standard deviation $\sigma_t(x_0,x_1)=\sigma\in [0,1]$, and a corresponding linear flow $x_t=\mu_t(x_0,x_1)+x\sigma_t(x_0,x_1)$ which is induced by  the conditional velocity vector $v_t(x|x_0,x_1)=x_1-x_0$. ICFM approximates the velocity vector by regressing a neural network $v^{\theta}_t$ against $x_1-x_0$ at each time $t\in [0,1]$, which lead to minimizing the following loss function:\begin{equation}
    \mathcal{L}_{ICFM}(\theta) = E_{t, p((x_0,x_1)) ,  p_t(x|(x_0,x_1))} \|v^{\theta}_t(x_t)-(x_1-x_0)\|^2 
    \label{ICFM}
\end{equation}
\subsubsection{Forest Flow (FF)}
% \textcolor{red}{To review}
Forest Flow is a ICFM-based  method (with $\sigma=0)$  developed by \cite{jolicoeurmartineau2024generating}  that makes use of a Xgboost regressor model \citep{chen2016xgboost}  $v^t_{\theta}$ to minimize the loss function $\mathcal{L}_{ICFM}$ defined in (Eq.(\ref{ICFM})).
Minimizing an unbiased estimate of  $\mathcal{L}_{ICFM}$ requires a random sampling from the original data $x_1$ (per batch sampling) through the empirical risk minimization (approximation of the expectation through average loss per dataset). 

In the case of neural network, the mini-batch stochastic gradient descent (SGD) framework requires  a training per batch which allows for the minimization of an  unbiased estimate of $\mathcal{L}_{ICFM}$. However, the Xgboost training  does not use the mini-batch training strategy, it rather uses  the whole data for training. Therefore, computing an unbiased estimate for $\mathcal{L}_{ICFM}$ in this case becomes challenging.  
However, because the  minimization of $\mathcal{L}_{ICFM}$ involves minimizing an expectation over all possible noise-data pairs ($x_0,x_1$), \cite{jolicoeurmartineau2024generating} chose to sample several Gaussian noisy data $x^i_0$ per data sample ($x_1$), with $x^i_0$ and $x_1$ having the same size (e.g. [d,n]).

Consequently, in case the number of sampled noisy data $z$ is $n_s$, this could be defined as a collection $C=\{(x^i_0, x_1)\mid i=0,\dots,n_s\}$ which is also equivalent to duplicating the data $x_1$ $n_s$ times and by this, creating a duplicated data $X_1$ and sample a Gaussian noise $X_0$  with size $[n_s\times d,n]$. The conditional flow is now defined as $X_t=(1-t)X_0+ tX_1$  at time  $t$ and used to minimize $\mathcal{L}_{ICFM}$ and then approximate the flow from $X_0$ to $X_1$ which is further  rescaled unto the initial data size $[d,n]$.
\section{Method} 
This section begins by elaborating on the \emph{continuous} Sequential Feature Forest Flow (CS3F) method, starting with its motivation and then explaining the method in detail. The second part of the section discusses the \emph{heterogeneous} Sequential Feature Forest Flow (HS3FM) method.
%\footnote{More details about the CS3F and HS3F can be respectively  found in the appendix section \ref{CS3Fatching} and \ref{HS3F}}
\subsection{\emph{Continuous} Sequential Feature Forest Flow (CS3F)}
\label{CS3F__}
Let us consider the initial original dataset $x_1=\{x^1,\dots,x^K\}\in \mathbb{R}^{n\times K}$ which has $K$ feature vectors, $D_K= \{1,\dots,K\}$ the set of indices of the features of $x_1$ and a set of standard Gaussian vectors $x_0=\{x^k_0,\dots,x^K_0\}\in \mathbb{R}^{n\times K}$.
In FF, an Xgboost regressor 
$v^t_{\theta}$ is trained to direct the trajectory of the flow $\hat{\phi}_t$ which primarily start from $x_0$ ($\hat{\phi}_{t=0}(x)=x_0$).

\begin{algorithm}[ht!]
% \footnotesize
\normalsize
\caption{Continuous Sequential Feature ForestFlow Algorithm}
\label{Algo1}
\KwIn{Data $x_1=\{x^1,\dots,x^K\}\in \mathbb{R}^{n\times K}$
(M continuous features and K-M categorical features), noise levels $n_s=50$, $t\in T_{level}$ ($T_{level}$ contains $n_s$ evenly spaced time values between 0 and 1).}
\For{$k \quad \text{in} \quad  [1,\dots,K]$}{  (Training process) 

\For{$t \quad \text{in} \quad  T_{level}$}{ 

\text{Sample} $x^k_{noise} \sim \mathcal{N}(0,I_n)$ and  compute conditional flow $x^k _t=(1-t)x^k_{noise} + tx^k$, and 
velocity vector $v^k _t=x^k-x^k_{noise}$.

Let   $X_{k-1}=(x^k _t,x^1,\dots, x^{k-1})$,  $X_{0}=x^1 _t$ and initialize  Xgboost regressor $f^{\theta^k _t}$

Solve $\underset{\theta^k _t}{argmin} L_2 \left(v^k _t,f^{\theta^k _t}(X_{k-1}) \right ).$ 
}
Store the  $n_s$ trained  models $\hat{f}^{\theta^k _t},\,  \text{for each} \, \,  k$}
\For{$k \quad \text{in} \quad  [1,\dots,K]$}{ (Feature Generation process)

\For{t in $T_{level}$}{ 
\text{Sample} $z^k_{t}\sim \mathcal{N}(0,I_n)$ and concatenate noisy vector with previously generated features $\tilde{X}_{k-1}=(z^k_{t},\Tilde{x}^{1}_1,\dots, \Tilde{x}^{k-1}_1)$, $\Tilde{X}_{0}=z^1_{t}$

Use $\Tilde{x}^k_{t_0}=z^{k}_{t}$ as initial condition 
for the Euler or Runge Kutta Discretization of: $\left( \dfrac{\partial \Tilde{x}^k_t}{\partial t} = \hat{f}^{\theta^k _t}(\tilde{X}_{k-1})\right)$}
We obtain the generated feature $\Tilde{x}^{k}_1$
}
The generated data $\tilde{X} = \{ \Tilde{x}^{1}_1,\dots,\Tilde{x}^{K}_1 \}$ is returned\
\Return{$\tilde{X}$}
\end{algorithm}

However, when the initial condition of the ODE (See Eq.(\ref{vct}))  slightly deviates from $x_0$,  $v^t_{\theta}$ does not provide useful directions to $\hat{\phi}_t$ which results in an accumulation of approximation errors while solving the flow ODE and therefore to an ill-approximation of the probability path. This leads to a significant increase in the Wasserstein distance value between the generated and real data distribution (see table (\ref{initcond})).

The \emph{Continuous} Sequential Feature Forest Flow (CS3F), is a FF approach developed to enhance the robustness of FF  to change in ODE flow initial condition $(\hat{\phi}_{t=0}(x)=x_0)$ by developing a per-feature autoregressive data generation which allows for the mitigation of the dependency on $(\hat{\phi}_{t=0}(x)=x_0)$. Furthermore, it also improves generation quality and the speed of FF.

Generating  features autoregressively through the FF strategy leads to training a different Xgboost model per feature $v^{\theta^k _t}$ (where $\theta^k _t$ is the per feature model parameter) that takes as input the per feature conditional flow $x^k_t=(1-t)x^k_0+tx^k$ combined  with the feature $x^1,\dots, x^{k-1}$, to learn the per feature velocity vector $v^k _t(x|x_0^k,x^k)=x^k-x^k_0$ for $k\in D_K$.

For each index $k\in D_K$ and for all time $t \in [0,1]$, we obtained a trained model  $\hat{v}^{\theta^k _t}$  that has learned $x^k-x^k_0$. Afterward, the flow for generating the $kth$ feature vector( $\Tilde{x}^k_1$) is determined by solving the following flow ODE $\dfrac{\partial \Tilde{x}^k_t}{\partial t}= \hat{v}^{\theta^k _t}( z^k_t| \Tilde{x}^{1}_1,\cdots,\Tilde{x}^{k-1}_1)$.

Where $\hat{v}^{\theta^k _t}$ takes as input a noisy vector $z^k_t\sim p^k_0$ given the previously generated features $\{\Tilde{x}^{1}_{t},\cdots,\Tilde{x}^{k-1}_{t}, \, t=1\}$. 
 Given this formulation, the only time that the vector field $\hat{v}^{\theta^k _t}$ provides the direction of its induced flow $\hat{\phi}^k_t$ only from a typical noise vector $z^k_t$ is when $k=1$ (first feature generation). Otherwise, $\hat{v}^{\theta^k _t}$ always directs $\hat{\phi}^k_t$  relying not only on $z^k_t$ but also on the filtered information provided by the previously generated features $\{\Tilde{x}^{1}_1,\cdots,\Tilde{x}^{k-1}_1\}$.

 This framework reduces the dependency on the ODE noisy initial conditions $v^k_{noise}$ and enhances model robustness to change in ODE noisy initial condition and provides faster results. Additionally, it can provide highly realistic continuous data. At the end of this autoregressive generation, we obtain the complete generated data $\hat{X}=
 \{\Tilde{x}^1_1,\dots,\Tilde{x}^K_1\}$ following a distribution similar to the original data $x_1$. 

\subsection{\emph{Heterogeneous} Sequential Feature Forest Flow (HS3F)}
% As $CS3F$ is deployed to handle continuous space, 
In this part, $CS3F$ is only for continuous feature generation.

Let us assume that the data set $x_1$ contains $N_c$  discrete features and let $I_{cat}$ and $I_{cont}$  be the disjoints respective set of indices of categorical and continuous features in $x_1$ ($ D_K= I_{cat}\bigcup I_{cont}$). Let us denote by $v_{cat}=\{x^{k}   \mid k\in I_{cat}\} \in x_1$, the set of all categorical features in $x_1$.

To generate discrete features $\Tilde{x}^k, \, k\in I_{cat}$ we first of all, respectively train different Xgboost classifiers $f^{\theta^k_t}$ to learn each  discrete features $x^k\in V_{cat}$ using the previous features $X_{k-1}=\{ x^{1},\dots,x^{k-1}\}$ as input. 
% \newline \newline 

  After training,  we use the previously generated features $\Tilde{X}_{k-1}=\{\Tilde{x}^{1},\dots, \Tilde{x}^{k-1}\}$ as a test input to the trained classifier $\hat{f}^{\theta^k_t}$ and predict the class probabilities for each $x^k\in V_{cat}$, which are used as a proportion for a multinomial class sampling to generate each discrete feature. 
  \begin{center}
\begin{algorithm}[ht]
% \small
\normalsize
    \caption{Heterogeneous Sequential FeatureForestFlow (HS3F) Algorithm }
    \label{Algo2}
    \KwIn{Data $x_1=\{x^1,\dots,x^K\}\in \mathbb{R}^{n\times K}$
($M$ continuous features, and $K-M$ categorical features), noise levels $n_s=50$, $t\in T_{level}$ ($T_{level}$ contains $n_s$ evenly spaced time values between 0 and 1).} 
\For{$k \quad \text{in} \quad  [1,\dots,K]$}{

\If{ $x^k$  is categorical}{  Get a Xgboost classifier model $f^{\theta^k}$
\\
Solve $\underset{\theta^k}{argmin} \,  Loss \left( x^k,f^{\theta^k}(x^1,\dots, x^{k-1})  \right )$ \\
Store the  trained classifier models $\hat{f}^{\theta^k}$ 
% Get a Xgboost regressor $f^{\theta^k _t}$ \\  
}\Else{ 
Use CS3F training process for  $f^{\theta^k_t}$ at the $n_s$ noise levels and 
store the $n_s$ trained regressor models $\hat{f}^{\theta^k _t}$ for each $k$,$t$
}
}  

\For{$k \quad \text{in} \quad  [1,\dots,K]$}{ 
\If{$x^k$ is categorical}{ 
Get the predicted class probabilities from
$\hat{f}^{\theta^k}(\Tilde{x}^{1}_1,\dots,\Tilde{x}^{k-1}_1).$
%=Prob\left[class1, \dots,classN\right]$ 
\\
$\Tilde{x}^k=Multinomial \left(\hat{f}^{\theta^k}( \Tilde{x}^{1}_1,\dots, \Tilde{x}^{k-1}_1) \right)$
}
\Else{
Use the CS3F generation process to get $\Tilde{x}^k=\Tilde{x}^k_1$
}
$\Tilde{x}^k$ is the generated feature.}

The generated data $\hat{X}=\{ \Tilde{x}^{1}_1,\dots, \Tilde{x}^{k}_1\}$ is returned \\
\Return{$\hat{X}$}
\end{algorithm}
\end{center}
  
  The \emph{Heterogeneous} Sequential Feature Forest Flow (HS3F)  uses  CS3F for continuous features generation ($\Tilde{x}^k, k\in I_{cont}$) and the multinomial sampling-based Xgboost classifier afore-mentioned for categorical features  ($\Tilde{x}^k, k\in I_{cat}$) sampling.

\section{ Experiments}
All experiments used an 8-core CPU machine with 16 GB RAM with Python 3.11.4 \citep{van1995python}. We use XGBoost as our prediction model, following \citet{jolicoeurmartineau2024generating}.

The Xgboost classifier used $\eta=0.1$ as learning rate and 115 total trees. All other hyperparameters for the Xgboost regression model and classifier were left are their default values.

As is done in \cite{jolicoeurmartineau2024generating}, we duplicated the rows of the dataset K = 100 times (note that this is unnecessary when the data set is entirely categorical) and we used $n_s=50$ noise levels. %(instead of the original $n_s=100$ in FF).

 We used conditional generation (generation conditioned to output labels for the classification dataset)\citep{NIPS2014_c60d060b,jolicoeurmartineau2024generating} to improve the performance of our methods (not necessary as it provides similar performances with unconditional generation). We min-max scaled the continuous variable (only) of each data before passing them into our models, and then we unscaled the generated data. 
 
 We applied the same processing as FF for the generated continuous features (min-max scaling and clipping after generation). For the experiments, we compared FF to CS3F and HS3F with Euler or Runge-Kutta 4th order solvers \citep{akinsola2023numerical}. This leads to 4 variants of our methods that are denoted as $HS3FM$-based Euler solver (HS3F-Euler), $HS3FM$-based  Runge-Kutta 4th order solver,  (HS3F-Rg4), $CS3F$-based Euler solver (CS3F-Euler)  and $CS3F$-based  Runge-Kutta 4th order solver (CS3F-Rg4).
% \noindent
\subsection{Assessment Metrics Description}
\begin{table*}[ht!]
  \centering
     \normalsize
    \caption{Metrics Information}
   \resizebox{1.0\textwidth}{!}{
  \hskip-1.0cm\begin{tabular}{|c|c|c|c|c|}
  
   \hline
  Metric& Notation& Description&Purpose&Used Package\\
   \hline 
  \cite{ruschendorf1985wasserstein}& $ W_{tr}$ and& Distance in distributions of& Distribution& POT\\
  Wasserstein1 distance $\downarrow$  & $W_{te} $ & ( $D^{fake}_{tr}$ vs  $D^{real}_{tr}$) and ($D^{fake}_{tr}$ vs $D^{real}_{te}$ )& Preservation&\citep{flamary2021pot}
    \\ 
    \hline 
   \cite{goutte2005probabilistic} &  $F1_{fake}$ and  $F1_{comb}$  & Classification  performances of models trained   & Usefulness for &Scikit-Learn
   \\
   F1 Score $\uparrow$& &on $D^{fake}_{tr}$  and $D^{Aug}_{tr}$ then tested on $D^{real}_{te}$&classification&  \citep{pedregosa2011scikit}\\  
   % \\&  & & \\ 
    \hline 
    \cite{ozer1985correlation} Coefficient of  & $R2_{fake}$ and $R2_{comb}$  &  Fit quality metric for regression model trained & Usefulness for &Scikit-Learn
   \\determination$\uparrow$ & &   on $D^{fake}_{tr}$ and $D^{Aug}_{tr}$    then  tested on $D^{real}_{te}$& regression& \citep{pedregosa2011scikit}\\ 
    % determination   &   & \\
    \hline 
    \cite{lohr2012coverage}Coverage $\uparrow$ & coverage$_{tr}$  and  & Measure of generation diversity ( $D^{fake}_{tr}$ vs  $D^{real}_{tr}$) &Sample diversity& Pandas, Numpy, Scikit-Learn  \\ & coverage$_{te}$&  and ($D^{fake}_{tr}$ vs $D^{real}_{te}$ )& & \citep{mckinney2010data,2020NumPy-Array}\\
     \hline 
    Running Time (in second)$\downarrow$ & time   & Determine the data generation process duration& Efficiency&\citep{python-time}\\
    \hline
  \end{tabular}
 }
  \label{tab:placeholder_label}
\end{table*}
\begin{table*}[ht!]
\centering
\normalsize
    \caption{Average post-generation metrics over 25 data sets (best in \textbf{bold})}
    \vspace{-0.5em}
    % \noindent
  \resizebox{1.0\textwidth}{!}{
  \hskip-0.35cm 
    \begin{tabular}{|c|cc|cc|cc|cc|c|}
    % \toprule 
    % \\
     \hline
     % \\
    Models& $W_{tr} \downarrow$ & $W_{te} \downarrow$  & $R2_{fake} \uparrow$ & $R2_{comb} \uparrow$ & $F1_{fake} \uparrow$  & $F1_{comb} \uparrow$ & $coverage_{tr} \uparrow$ & $coverage_{te} \uparrow$& time (sec)$\downarrow$\\
    % \midrule
     \hline
    HS3F-Euler & 1.283 & 1.949 & 0.580 & 0.668 & 0.738 & \textbf{0.775} & 0.787 & 0.846 & 278.425\\
    CS3F-Euler & 1.349 & 1.980 & 0.580 & 0.668 & 0.724 & 0.766 & 0.783 & 0.863 & \textbf{262.556} \\
    HS3F-Rg4 &\textbf{1.233} & 1.903 & 0.592 &\textbf{0.672} &  \textbf{0.741} & 0.773 & 0.819 & 0.861 & 331.058\\
    CS3F-Rg4 & 1.405 & 1.997 & 0.592 & \textbf{0.672} & 0.712 & 0.760 & 0.786 & 0.867 & 332.877   \\
    ForestFlow & 1.356 & \textbf{1.898} & \textbf{0.606} & 0.659 &  0.723 & 0.766 & \textbf{0.839}& \textbf{0.894} & 1073.766 \\
    % Stasy & 3.409244 & 3.660540 & 0.664667 & \textbf{0.654377} & 0.639889 & 0.771394 & 0.614322 \\
    % TabDDM & 4.271244 & 4.787619 & 0.664667 & 0.605101 & 0.660409 & 0.771394 & 0.649011 
    \hline
    % \bottomrule
\end{tabular} 
}
 \label{tab:performance_metrics_1}
\end{table*}

The original data set is divided into train ($D^{real}_{tr}$) and test data ($D^{real}_{te}$) and all model generation is based on $D^{real}_{tr}$. 
The generated data are denoted by $D^{fake}_{tr}$ and $D^{real}_{tr}$ augmented with $D^{fake}_{tr}$  will be denoted by $D^{Aug}_{tr}$. 

 We averaged the F1 score and the R-squared results of four classifiers/ regressors based on Adaboost, Random Forest, Xgboost and Logistic Regression trained on $D^{fake}_{tr}$ and  $D^{Aug}_{tr}$ which resulted in two respective F1 scores ($F1_{fake}$ and $F1_{comb}$) and two respective R2 scores ($R2_{fake}$ and $R2_{comb}$).

   We also determined the respective Wasserstein 1 distances (with L1 norm) between $D^{real}_{tr}$ to $D^{fake}_{tr}$ and from $D^{real}_{tr}$  to $D^{real}_{te}$  denoted by $W_{tr}$ and $W_{te}$. We also determined the respective coverage indicator (coverage$_{tr}$ and coverage$_{te}$) of $D^{fake}_{tr}$ with respect to $D^{real}_{te}$ and $D^{fake}_{tr}$. Table (\ref{tab:placeholder_label}) provides complementary information.

\subsection{Comparison of Model Generation Using Default ODE Initialization in CFM (Standard gaussian)}
We evaluated our methods on 25 real-world datasets from the UCI Machine Learning Repository (\url{https://archive.ics.uci.edu/datasets}) and scikit-Learn \citep{pedregosa2011scikit} .  Among these 25 data sets,  6  have continuous output, while the remaining ones (19) have categorical outputs.
We compare the results of our four models to Forest Flow default approach \citep{jolicoeurmartineau2024generating} using all the metrics described in table (\ref{tab:placeholder_label}).
% Stasy \citep{kim2023stasy},  TabDDPM\citep{10.5555/3618408.3619133}.
To condensate the information, we average the result of each method over the 25 datasets and 
gather them in Table  
(\ref{tab:performance_metrics_1}).  To evaluate the performance of our methods on data having categorical features, we used the data sets among the 25 that have at least 20\% of categorical features and we found five of them that are blood\_transfusion\citep{yeh2008blood}, congress \citep{congressional_voting}, car \citep{bohanec1997car}, tictactoe \citep{aha1991tic_tac_toe}, and glass \citep{german1987glass}. For comparison purposes, we recorded the performances of our methods along with the performance of   ForestFlow on these data sets in table (\ref{tab:performance_metrics_2}).  
% We provided more information about individual model performances per data in the appendix (subsection \ref{barplotss}).
\footnote{ More information about the data sets and the individual model performances per data set are provided in the appendix (subsection (\ref{UCI}) and (\ref{barplotss})).}
% , the comparison metrics plots across all the datasets used and  for all methods used in this experiment .
\begin{table*}[ht!]
    \centering
      \caption{Average post-generation metrics over five datasets (among the 25 used previously) blood\_transfusion\citep{yeh2008blood}, congress \citep{congressional_voting}, car \citep{bohanec1997car}, tictactoe \citep{tic_tac_toe}, glass \citep{german1987glass} having at least 20\% of categorical variables (best in \textbf{bold}).} \vspace{-0.5em}
    \begin{tabular}{|c|cc|cc|cc|c|}
    % \toprule
    % \\
    \hline
    Models& $W_{tr} \downarrow$ & $W_{te} \downarrow$ & $F1_{fake} \uparrow$ & $F1_{comb} \uparrow$ & $coverage_{tr} \uparrow$ & $coverage_{te} \uparrow$ & time (sec) $\downarrow$ 
    \\
\hline 
    HS3F-Euler & 0.596 & 1.321 & \textbf{0.763} & \textbf{0.787} & 0.788 & 0.671 & \textbf{17.612} \\
    CS3F-Euler & 0.926 & 1.473 & 0.709 & 0.755 & 0.771 & \textbf{0.756} & 49.186 \\
    HS3F-Rg4   & \textbf{0.584} & \textbf{1.313} & 0.747 & 0.756 & \textbf{0.804} & 0.661 & 22.494 \\
    CS3F-Rg4   & 1.448 & 1.780 & 0.637 & 0.707 & 0.636 & 0.692 & 66.601 \\
    ForestFlow  & 1.064 & 1.461 & 0.703 & 0.747 & 0.700 & 0.735 & 468.352 
    \\
    % \bottomrule
    \hline
    \end{tabular}
     \label{tab:performance_metrics_2}
\end{table*}

\begin{table*}[ht!]
\centering
\caption{Sensitivity to ODE initial condition with $\Delta W=\mid W^{modified}-W^{default}\mid$ (best in \textbf{bold}).}
\begin{tabular}{|c|ccc|ccc|}
\hline
\multirow{2}{*}{Initial Conditions } & \multicolumn{3}{c|}{$\Delta W_{tr}$ ($\downarrow$)} & \multicolumn{3}{c|}{$ \Delta W_{te}$ ($\downarrow$)} \\ \cline{2-7} 
                          & HS3F-Rg4 & CS3F-Rg4 &ForestFlow& HS3F-Rg4 & CS3F-Rg4 &ForestFlow  \\ \hline
$\mathcal{N}(0.1,(1.1)^2*I_n)$& 0.0085    & \textbf{0.0001}     & 0.1462   & 0.02148 &  \textbf{0.0001}    &0.1565   
\\
\hline
%$\mathcal{N}(0.0,(0.5)^2*I_n)$  & 0.007    &  cs3  & ff   & 0.006  & cs3   &  ff\\\hline 
$\mathcal{N}(0.0,(0.9)^2*I_n)$                    & 0.0018 &  \textbf{0.0006}   &0.03     &0.0163  & \textbf{0.0007}   &0.049     \\ \hline

%$\mathcal{N}(0.0,(0.5)^2*I_n)$  & 0.007    &  cs3  & ff   & 0.006  & cs3   &  ff\\\hline 
$\mathcal{N}(0.0,(1.1)^2*I_n)$                    & 0.0028 &  \textbf{0.0007}  &0.051     &0.0154 & \textbf{0.0008}  &0.0497    \\ \hline
\end{tabular}
\label{initcond}
\end{table*}

From table (\ref{tab:performance_metrics_1}), we can observe that   HS3F-Rg4 outperformed the other methods across $W_{tr}$, $R2_{comb}$ and $F1_{fake}$,
 showcasing its ability to provide distribution-preserving generated samples through relatively more minor Wasserstein distances.  It also demonstrates its usefulness for post-generation predictive experiments in regression and classification experiments through the R-squared and F1 scores. Across the other metrics, HS3F-Rg4 generally is the 2nd best method.
HS3F-Euler  closely follows the performances of HS3F-Rg4. Forestflow outperforms the other methods in terms of $W_{te}$, $coverage_{tr}$ and $coverage_{te}$ and $R2_{fake}$ indicating its ability to provide distribution preserving samples and can generalize better than the other methods. However, its performance in the remaining metrics is mostly ranked as the fourth-best method. We can also see that CS3F-Rg4 provided the worst overall performance in this experiment.  The S3F-based Euler solver methods generate data faster than S3F-based Rg4 solver methods, which are significantly faster than Forest Flow (3.2  to 4.1 times faster).

 Table (\ref{tab:performance_metrics_2}) shows that for data sets with at least 20\% of categorical features, HS3F-based models significantly outperform the other methods while delivering the fastest generation time. 
 They generate data 20.82 to 26.59 times faster than ForestFlow, and their overall performance is followed by that of CS3F-based Euler solver and then Forestflow. Moreover, CS3F-Rg4 still achieves the poorest overall performance.

An intriguing insight from this experiment is the role of the CS3F-based  models. While they may perform poorly when evaluated in isolation, they proved to be highly effective within the HS3F framework. Specifically, CS3F-Rg4 and CS3F-Euler contribute valuable, nuanced information that enhances the performance of Xgboost classifiers in this context. This fact is noticeable by comparing the performances of CS3F-Rg4 to  HS3F-Rg4. This suggests that, despite its limitations, CS3F-Rg4 offers unique features that enrich the prediction for the Xgboost Classifiers predictions in HS3F, leading to improved discrete feature generation.
\subsection{Wasserstein Distance Sensitivity to Affine Transformation In ODE Initialization  } \label{comp_drop}
As discussed in the section (\ref{CS3F__}), in the context of FF, when the velocity vector field is trained on a noisy data $x_0$ whose distribution differs from that of the flow ODE initial condition noisy data $(\phi_{t=0})$, the resulting flow is likely to have accumulated approximation errors that will result in an ill-approximation of the probability path and therefore lead to higher Wasserstein distance between the real and approximated data distribution (higher $W_{tr}$ and $W_{te}$).

To confirm our theory, we slightly modify the data distribution of the ODE flow initial condition using an affine transformation of the initial noisy data distribution $x_0$ (Standard Gaussian) leading to $\phi_{t=0}
\sim \mathcal{N}(\mu=(0.1,\dots,0.1),\sigma^2=(1.1)^2*I_n)$, $\phi_{t=0}
\sim \mathcal{N}(\mu=(0.0,\dots,0.),\sigma^2=(0.9)^2*I_n)$  and $\phi_{t=0}
\sim \mathcal{N}(\mu=(0.0,\dots,0.),\sigma^2=(1.1)^2*I_n)$. After that, we averaged the obtained Wasserstein train and test of Forest flow, CS3F-Rg4, and HS3F-Rg4 over the 25 data sets (aforementioned) and determined the Wasserstance distance difference between the modified and default initial conditions results in absolute values. The results are recorded in table (\ref{initcond}) and compare them to the results obtained using the default initial condition (see table (\ref{tab:performance_metrics_1})).

The results of the experiment, as shown in table (\ref{initcond}), confirm our theory showing that CS3F-Rg4 provides the most robust Wasserstein distance performance which is followed by HS3F-Rg4 and then comes ForestFlow which appears to be significantly impacted by the change. 
%It should be noted that HS3F-Rg4 provided the best Wasserstein train and test.
\section{Conclusion}
%Restate the Purpose

In this work, we extended Forest Flow by addressing its existing limitations which are sensitivity to initial condition change and lack of efficient strategy to handle mixed type data.

%Summarize Key Findings
Our experimental results confirmed that, compared to Forest Flow, HS3F yields methods that are more robust to affine change in the initial condition of the ODE flow and provides a more explicit and efficient way to handle mixed data types while improving overall data generation performance across several data sets and providing faster results. 

% Potential
This framework offers a new perspective on data generation using diffusion-based techniques, emphasizing the sequential generation of features and the utilization of information from previously generated features. HS3F represents a significant step forward in the field of data generation, offering new opportunities to improve the efficiency of generative models.
%#Future work
Spurious features may negatively impact sequential generation. A helpful practice would be to determine which features are causally related and use only those in the sequential generation process.
% Another research direction is to integrate the S3F strategy into other generative models.
% Strong statement of end
% In general, \\newpage
\section*{Acknowledgments}
This research was enabled in part by compute resources provided by Mila (MILA-Quebec AI Institute ). A.-C. Akazan acknowledges funding from the IVADO-AIMS program during his research internship program. 

%\newpage
\normalsize
\bibliographystyle{abbrvnat}
\bibliography{ref.bib} 

\begin{thebibliography}{59}
\providecommand{\natexlab}[1]{#1}
\providecommand{\url}[1]{\texttt{#1}}
\expandafter\ifx\csname urlstyle\endcsname\relax
  \providecommand{\doi}[1]{doi: #1}\else
  \providecommand{\doi}{doi: \begingroup \urlstyle{rm}\Url}\fi

\bibitem[Aeberhard and Forina(1991)]{aeberhard1991wine}
S.~Aeberhard and M.~Forina.
\newblock Comparison of methods for classifying between 3 varieties of wine.
\newblock \emph{Data in Brief}, 1991.

\bibitem[Aha(1991)]{aha1991tic_tac_toe}
D.~W. Aha.
\newblock Comparing case-based reasoning and other machine learning techniques in tic-tac-toe dataset.
\newblock \emph{Machine Learning Journal}, 1991.

\bibitem[Akinsola(2023)]{akinsola2023numerical}
V.~Akinsola.
\newblock Numerical methods: Euler and runge-kutta.
\newblock In \emph{Qualitative and Computational Aspects of Dynamical Systems}. IntechOpen, 2023.

\bibitem[Bhatt(2012)]{bhatt2012planning_relax}
P.~Bhatt.
\newblock A planning relaxation approach for temporal networks.
\newblock In \emph{ICAPS Proceedings}, 2012.

\bibitem[Bohanec(1997)]{bohanec1997car}
M.~Bohanec.
\newblock Car evaluation.
\newblock \emph{UCI Machine Learning Repository}, 10:\penalty0 C5JP48, 1997.

\bibitem[Borisov et~al.(2024)Borisov, Leemann, Seßler, Haug, Pawelczyk, and Kasneci]{9998482}
V.~Borisov, T.~Leemann, K.~Seßler, J.~Haug, M.~Pawelczyk, and G.~Kasneci.
\newblock Deep neural networks and tabular data: A survey.
\newblock \emph{IEEE Transactions on Neural Networks and Learning Systems}, 35\penalty0 (6):\penalty0 7499--7519, 2024.
\newblock \doi{10.1109/TNNLS.2022.3229161}.

\bibitem[Brooks et~al.(2014)Brooks, Pope, and Marcolini]{brooks2014airfoil}
T.~F. Brooks, D.~S. Pope, and M.~A. Marcolini.
\newblock Airfoil self-noise and prediction.
\newblock \emph{NASA Technical Report}, 2014.

\bibitem[Charytanowicz(2012)]{charytanowicz2012seeds}
M.~Charytanowicz.
\newblock Assessment of k-means clustering with genetic algorithms for seed dataset.
\newblock \emph{Scientific Research Journal}, 2012.

\bibitem[Chen and Guestrin(2016)]{chen2016xgboost}
T.~Chen and C.~Guestrin.
\newblock Xgboost: A scalable tree boosting system.
\newblock In \emph{Proceedings of the 22nd ACM SIGKDD international conference on knowledge discovery and data mining}, pages 785--794. ACM, 2016.

\bibitem[Cortez et~al.(2009)]{cortez2009wine_quality}
P.~Cortez et~al.
\newblock Modeling wine preferences by data mining from physicochemical properties.
\newblock \emph{Decision Support Systems}, 2009.

\bibitem[Deterding(1997)]{deterding_vowel}
P.~Deterding.
\newblock Speaker classification in a vowel recognition benchmark.
\newblock 1997.

\bibitem[Dias and et~al.(2009)]{dias2009libras}
J.~Dias and et~al.
\newblock Libras gesture classification: comparing svm, knn and others.
\newblock \emph{Pattern Recognition Letters}, 2009.

\bibitem[Esteban et~al.(2017)Esteban, Hyland, and Rätsch]{esteban2017realvalued}
C.~Esteban, S.~L. Hyland, and G.~Rätsch.
\newblock Real-valued (medical) time series generation with recurrent conditional gans, 2017.

\bibitem[Fisher(1988)]{fisher1988iris}
R.~Fisher.
\newblock The use of multiple measurements in taxonomic problems.
\newblock \emph{Annals of Eugenics}, 1988.

\bibitem[Flamary et~al.(2021)Flamary, Courty, Gramfort, Alaya, Boisbunon, Chambon, Chapel, Corenflos, Fatras, Fournier, Gautheron, Gayraud, Janati, Rakotomamonjy, Redko, Rolet, Schutz, Seguy, Sutherland, Tavenard, Tong, and Vayer]{flamary2021pot}
R.~Flamary, N.~Courty, A.~Gramfort, M.~Z. Alaya, A.~Boisbunon, S.~Chambon, L.~Chapel, A.~Corenflos, K.~Fatras, N.~Fournier, L.~Gautheron, N.~T. Gayraud, H.~Janati, A.~Rakotomamonjy, I.~Redko, A.~Rolet, A.~Schutz, V.~Seguy, D.~J. Sutherland, R.~Tavenard, A.~Tong, and T.~Vayer.
\newblock Pot: Python optimal transport.
\newblock \emph{Journal of Machine Learning Research}, 22\penalty0 (78):\penalty0 1--8, 2021.
\newblock URL \url{http://jmlr.org/papers/v22/20-451.html}.

\bibitem[German and Spiehler(1987)]{german1987glass}
B.~German and V.~Spiehler.
\newblock Glass identification database.
\newblock \emph{UC Irvine Machine Learning Repository}, 1987.

\bibitem[Gerritsma and et~al.(2013)]{gerritsma2013yacht}
M.~Gerritsma and et~al.
\newblock Yacht resistance predictions using machine learning.
\newblock \emph{Journal of Fluid Mechanics}, 2013.

\bibitem[Goodfellow et~al.(2014)Goodfellow, Pouget-Abadie, Mirza, Xu, Warde-Farley, Ozair, Courville, and Bengio]{goodfellow2014generative}
I.~Goodfellow, J.~Pouget-Abadie, M.~Mirza, B.~Xu, D.~Warde-Farley, S.~Ozair, A.~Courville, and Y.~Bengio.
\newblock Generative adversarial nets.
\newblock In \emph{Advances in neural information processing systems}, pages 2672--2680, 2014.

\bibitem[Goutte and Gaussier(2005)]{goutte2005probabilistic}
C.~Goutte and E.~Gaussier.
\newblock A probabilistic interpretation of precision, recall and f-score, with implication for evaluation.
\newblock In \emph{European conference on information retrieval}, pages 345--359. Springer, 2005.

\bibitem[Harris et~al.(2020)Harris, Millman, van~der Walt, Gommers, Virtanen, Cournapeau, Wieser, Taylor, Berg, Smith, Kern, Picus, Hoyer, van Kerkwijk, Brett, Haldane, Fernández~del Río, Wiebe, Peterson, Gérard-Marchant, Sheppard, Reddy, Weckesser, Abbasi, Gohlke, and Oliphant]{2020NumPy-Array}
C.~R. Harris, K.~J. Millman, S.~J. van~der Walt, R.~Gommers, P.~Virtanen, D.~Cournapeau, E.~Wieser, J.~Taylor, S.~Berg, N.~J. Smith, R.~Kern, M.~Picus, S.~Hoyer, M.~H. van Kerkwijk, M.~Brett, A.~Haldane, J.~Fernández~del Río, M.~Wiebe, P.~Peterson, P.~Gérard-Marchant, K.~Sheppard, T.~Reddy, W.~Weckesser, H.~Abbasi, C.~Gohlke, and T.~E. Oliphant.
\newblock Array programming with {NumPy}.
\newblock \emph{Nature}, 585:\penalty0 357–362, 2020.
\newblock \doi{10.1038/s41586-020-2649-2}.

\bibitem[Ho et~al.(2020)Ho, Jain, and Abbeel]{NEURIPS2020_4c5bcfec}
J.~Ho, A.~Jain, and P.~Abbeel.
\newblock Denoising diffusion probabilistic models.
\newblock In H.~Larochelle, M.~Ranzato, R.~Hadsell, M.~Balcan, and H.~Lin, editors, \emph{Advances in Neural Information Processing Systems}, volume~33, pages 6840--6851. Curran Associates, Inc., 2020.
\newblock URL \url{https://proceedings.neurips.cc/paper_files/paper/2020/file/4c5bcfec8584af0d967f1ab10179ca4b-Paper.pdf}.

\bibitem[Ishfaq et~al.(2023)Ishfaq, Hoogi, and Rubin]{ishfaq2023tvae}
H.~Ishfaq, A.~Hoogi, and D.~Rubin.
\newblock Tvae: Triplet-based variational autoencoder using metric learning, 2023.

\bibitem[Jiang et~al.(2014)Jiang, Meng, Yu, Lan, Shan, and Hauptmann]{NIPS2014_c60d060b}
L.~Jiang, D.~Meng, S.-I. Yu, Z.~Lan, S.~Shan, and A.~Hauptmann.
\newblock Self-paced learning with diversity.
\newblock In Z.~Ghahramani, M.~Welling, C.~Cortes, N.~Lawrence, and K.~Weinberger, editors, \emph{Advances in Neural Information Processing Systems}, volume~27. Curran Associates, Inc., 2014.
\newblock URL \url{https://proceedings.neurips.cc/paper_files/paper/2014/file/c60d060b946d6dd6145dcbad5c4ccf6f-Paper.pdf}.

\bibitem[Jolicoeur-Martineau et~al.(2024)Jolicoeur-Martineau, Fatras, and Kachman]{jolicoeurmartineau2024generating}
A.~Jolicoeur-Martineau, K.~Fatras, and T.~Kachman.
\newblock Generating and imputing tabular data via diffusion and flow-based gradient-boosted trees, 2024.

\bibitem[Juwara et~al.(2024)Juwara, El-Hussuna, and {El Emam}]{JUWARA2024100946}
L.~Juwara, A.~El-Hussuna, and K.~{El Emam}.
\newblock An evaluation of synthetic data augmentation for mitigating covariate bias in health data.
\newblock \emph{Patterns}, 5\penalty0 (4):\penalty0 100946, 2024.
\newblock ISSN 2666-3899.
\newblock \doi{https://doi.org/10.1016/j.patter.2024.100946}.
\newblock URL \url{https://www.sciencedirect.com/science/article/pii/S266638992400045X}.

\bibitem[Kim et~al.(2021)Kim, Jeon, Lee, Hyeong, and Park]{kim2021octgan}
J.~Kim, J.~Jeon, J.~Lee, J.~Hyeong, and N.~Park.
\newblock Oct-gan: Neural ode-based conditional tabular gans, 2021.

\bibitem[Kim et~al.(2023)Kim, Lee, and Park]{kim2023stasy}
J.~Kim, C.~Lee, and N.~Park.
\newblock Stasy: Score-based tabular data synthesis, 2023.

\bibitem[Kingma and Welling(2013)]{kingma2013auto}
D.~P. Kingma and M.~Welling.
\newblock Auto-encoding variational bayes.
\newblock \emph{arXiv preprint arXiv:1312.6114}, 2013.

\bibitem[Kotelnikov et~al.(2023)Kotelnikov, Baranchuk, Rubachev, and Babenko]{10.5555/3618408.3619133}
A.~Kotelnikov, D.~Baranchuk, I.~Rubachev, and A.~Babenko.
\newblock Tabddpm: modelling tabular data with diffusion models.
\newblock In \emph{Proceedings of the 40th International Conference on Machine Learning}, ICML'23. JMLR.org, 2023.

\bibitem[LEE et~al.(2021)LEE, Hyeong, Jeon, Park, and Cho]{NEURIPS2021_22456f4b}
J.~LEE, J.~Hyeong, J.~Jeon, N.~Park, and J.~Cho.
\newblock Invertible tabular gans: Killing two birds with one stone for tabular data synthesis.
\newblock In M.~Ranzato, A.~Beygelzimer, Y.~Dauphin, P.~Liang, and J.~W. Vaughan, editors, \emph{Advances in Neural Information Processing Systems}, volume~34, pages 4263--4273. Curran Associates, Inc., 2021.
\newblock URL \url{https://proceedings.neurips.cc/paper_files/paper/2021/file/22456f4b545572855c766df5eefc9832-Paper.pdf}.

\bibitem[Lipman et~al.(2023)Lipman, Chen, Ben-Hamu, Nickel, and Le]{lipman2023flow}
Y.~Lipman, R.~T.~Q. Chen, H.~Ben-Hamu, M.~Nickel, and M.~Le.
\newblock Flow matching for generative modeling, 2023.

\bibitem[Little(2008)]{little2008parkinsons}
M.~Little.
\newblock Exploiting non-linear recurrence and fractal scaling properties for voice disorder detection.
\newblock \emph{Biomedical Engineering}, 2008.

\bibitem[Lohr(2012)]{lohr2012coverage}
S.~L. Lohr.
\newblock Coverage and sampling.
\newblock In \emph{International handbook of survey methodology}, pages 97--112. Routledge, 2012.

\bibitem[Lucas and et~al.(2013)]{lucas2013climate_model_crashes}
C.~Lucas and et~al.
\newblock Investigating crashes in climate models using machine learning.
\newblock \emph{Journal of Climate}, 2013.

\bibitem[Mansouri et~al.(2013)]{mansouri2013qsar_biodegradation}
K.~Mansouri et~al.
\newblock Qsar models for biodegradation of organic chemicals.
\newblock \emph{Chemical Research in Toxicology}, 2013.

\bibitem[McKinney et~al.(2010)]{mckinney2010data}
W.~McKinney et~al.
\newblock Data structures for statistical computing in python.
\newblock In \emph{Proceedings of the 9th Python in Science Conference}, volume 445, pages 51--56. Austin, TX, 2010.

\bibitem[Motamed et~al.(2021)Motamed, Rogalla, and Khalvati]{MOTAMED2021100779}
S.~Motamed, P.~Rogalla, and F.~Khalvati.
\newblock Data augmentation using generative adversarial networks (gans) for gan-based detection of pneumonia and covid-19 in chest x-ray images.
\newblock \emph{Informatics in Medicine Unlocked}, 27:\penalty0 100779, 2021.
\newblock ISSN 2352-9148.
\newblock \doi{https://doi.org/10.1016/j.imu.2021.100779}.
\newblock URL \url{https://www.sciencedirect.com/science/article/pii/S2352914821002501}.

\bibitem[Nakai(1996)]{nakai1996a}
K.~Nakai.
\newblock Prediction of protein cellular locations using genetic algorithms and neural networks.
\newblock \emph{Bioinformatics}, 12, 1996.

\bibitem[Ozer(1985)]{ozer1985correlation}
D.~J. Ozer.
\newblock Correlation and the coefficient of determination.
\newblock \emph{Psychological bulletin}, 97\penalty0 (2):\penalty0 307, 1985.

\bibitem[Park et~al.(2018)Park, Mohammadi, Gorde, Jajodia, Park, and Kim]{Park_2018}
N.~Park, M.~Mohammadi, K.~Gorde, S.~Jajodia, H.~Park, and Y.~Kim.
\newblock Data synthesis based on generative adversarial networks.
\newblock \emph{Proceedings of the VLDB Endowment}, 11\penalty0 (10):\penalty0 1071–1083, June 2018.
\newblock ISSN 2150-8097.
\newblock \doi{10.14778/3231751.3231757}.
\newblock URL \url{http://dx.doi.org/10.14778/3231751.3231757}.

\bibitem[Pedregosa et~al.(2011)Pedregosa, Varoquaux, Gramfort, Michel, Thirion, Grisel, Blondel, Prettenhofer, Weiss, Dubourg, et~al.]{pedregosa2011scikit}
F.~Pedregosa, G.~Varoquaux, A.~Gramfort, V.~Michel, B.~Thirion, O.~Grisel, M.~Blondel, P.~Prettenhofer, R.~Weiss, V.~Dubourg, et~al.
\newblock Scikit-learn: Machine learning in python.
\newblock \emph{Journal of machine learning research}, 12\penalty0 (Oct):\penalty0 2825--2830, 2011.

\bibitem[{Python Software Foundation}(2024)]{python-time}
{Python Software Foundation}.
\newblock \emph{time — Time access and conversions}, 2024.
\newblock URL \url{https://docs.python.org/3/library/time.html}.

\bibitem[R{\"u}schendorf(1985)]{ruschendorf1985wasserstein}
L.~R{\"u}schendorf.
\newblock The wasserstein distance and approximation theorems.
\newblock \emph{Probability Theory and Related Fields}, 70\penalty0 (1):\penalty0 117--129, 1985.

\bibitem[Sejnowski and Gorman()]{sejnowski_sonar}
T.~J. Sejnowski and P.~J. Gorman.
\newblock Parallel networks that learn to pronounce english text.
\newblock In \emph{IEEE Transactions on Neural Networks}.

\bibitem[Sigillito et~al.(1989)Sigillito, Wing, Hutton, and et~al.]{sigillito1989ionosphere}
V.~G. Sigillito, S.~Wing, J.~Hutton, and et~al.
\newblock Classification of radar returns from the ionosphere.
\newblock \emph{Johns Hopkins APL Technical Digest}, 1989.

\bibitem[Sohl-Dickstein et~al.(2015)Sohl-Dickstein, Weiss, Maheswaranathan, and Ganguli]{pmlr-v37-sohl-dickstein15}
J.~Sohl-Dickstein, E.~Weiss, N.~Maheswaranathan, and S.~Ganguli.
\newblock Deep unsupervised learning using nonequilibrium thermodynamics.
\newblock In F.~Bach and D.~Blei, editors, \emph{Proceedings of the 32nd International Conference on Machine Learning}, volume~37 of \emph{Proceedings of Machine Learning Research}, pages 2256--2265, Lille, France, 07--09 Jul 2015. PMLR.
\newblock URL \url{https://proceedings.mlr.press/v37/sohl-dickstein15.html}.

\bibitem[Song et~al.(2021)Song, Sohl-Dickstein, Kingma, Kumar, Ermon, and Poole]{song2021scorebasedgenerativemodelingstochastic}
Y.~Song, J.~Sohl-Dickstein, D.~P. Kingma, A.~Kumar, S.~Ermon, and B.~Poole.
\newblock Score-based generative modeling through stochastic differential equations, 2021.
\newblock URL \url{https://arxiv.org/abs/2011.13456}.

\bibitem[Tong et~al.(2023)Tong, Malkin, Huguet, Zhang, Rector-Brooks, Fatras, Wolf, and Bengio]{tong2023improving}
A.~Tong, N.~Malkin, G.~Huguet, Y.~Zhang, J.~Rector-Brooks, K.~Fatras, G.~Wolf, and Y.~Bengio.
\newblock Improving and generalizing flow-based generative models with minibatch optimal transport, 2023.

\bibitem[{UCI Machine Learning Repository}(1987)]{congressional_voting}
{UCI Machine Learning Repository}.
\newblock Congressional voting records data set.
\newblock \url{https://archive.ics.uci.edu/ml/datasets/Congressional+Voting+Records}, 1987.

\bibitem[{UCI Machine Learning Repository}(1991)]{tic_tac_toe}
{UCI Machine Learning Repository}.
\newblock Tic-tac-toe endgame data set.
\newblock \url{https://archive.ics.uci.edu/ml/datasets/Tic-Tac-Toe+Endgame}, 1991.
\newblock Accessed: 2024-09-03.

\bibitem[Van~Rossum and Drake~Jr(1995)]{van1995python}
G.~Van~Rossum and F.~L. Drake~Jr.
\newblock \emph{Python reference manual}.
\newblock Centrum voor Wiskunde en Informatica Amsterdam, 1995.

\bibitem[Wolberg et~al.(1995)Wolberg, Street, and Mangasarian]{wolberg1995breast_cancer}
W.~H. Wolberg, W.~N. Street, and O.~Mangasarian.
\newblock Breast cancer diagnosis and prognosis via linear programming.
\newblock \emph{Operations Research}, 1995.

\bibitem[Xu et~al.(2019)Xu, Skoularidou, Cuesta-Infante, and Veeramachaneni]{NEURIPS2019__254ed7d2}
L.~Xu, M.~Skoularidou, A.~Cuesta-Infante, and K.~Veeramachaneni.
\newblock Modeling tabular data using conditional gan.
\newblock In H.~Wallach, H.~Larochelle, A.~Beygelzimer, F.~d\textquotesingle Alch\'{e}-Buc, E.~Fox, and R.~Garnett, editors, \emph{Advances in Neural Information Processing Systems}, volume~32. Curran Associates, Inc., 2019.
\newblock URL \url{https://proceedings.neurips.cc/paper_files/paper/2019/file/254ed7d2de3b23ab10936522dd547b78-Paper.pdf}.

\bibitem[Yeh(2007)]{yeh2007concrete_compression}
I.-C. Yeh.
\newblock Modeling of strength of high-performance concrete.
\newblock \emph{Cement and Concrete Research}, 34:\penalty0 1429--1437, 2007.

\bibitem[Yeh(2008{\natexlab{a}})]{yeh2008blood}
I.-C. Yeh.
\newblock Blood transfusion service center.
\newblock \emph{UCI Machine Learning Repository}, 2008{\natexlab{a}}.

\bibitem[Yeh(2008{\natexlab{b}})]{yeh2008blood_transfusion}
I.-C. Yeh.
\newblock Modeling of blood donation system using machine learning algorithms.
\newblock \emph{Expert Systems with Applications}, 34:\penalty0 500--507, 2008{\natexlab{b}}.

\bibitem[Yeh(2009)]{yeh2009concrete_slump}
I.-C. Yeh.
\newblock Concrete slump test prediction using machine learning.
\newblock \emph{Journal of Materials in Civil Engineering}, 21:\penalty0 151--158, 2009.

\bibitem[Yoon et~al.(2019)Yoon, Jordon, and van~der Schaar]{yoon2018pategan}
J.~Yoon, J.~Jordon, and M.~van~der Schaar.
\newblock {PATE}-{GAN}: Generating synthetic data with differential privacy guarantees.
\newblock In \emph{International Conference on Learning Representations}, 2019.
\newblock URL \url{https://openreview.net/forum?id=S1zk9iRqF7}.

\bibitem[Zhao et~al.(2021)Zhao, Kunar, Birke, and Chen]{pmlr-v157-zhao21a}
Z.~Zhao, A.~Kunar, R.~Birke, and L.~Y. Chen.
\newblock Ctab-gan: Effective table data synthesizing.
\newblock In V.~N. Balasubramanian and I.~Tsang, editors, \emph{Proceedings of The 13th Asian Conference on Machine Learning}, volume 157 of \emph{Proceedings of Machine Learning Research}, pages 97--112. PMLR, 17--19 Nov 2021.
\newblock URL \url{https://proceedings.mlr.press/v157/zhao21a.html}.

\end{thebibliography}
% \documentclass{article}
% \usepackage{aistats2025}
% If your paper is accepted, change the options for the package
% aistats2025 as follows:
%
%\usepackage[accepted]{aistats2025}
%
% This option will print headings for the title of your paper and
% headings for the authors names, plus a copyright note at the end of
% the first column of the first page.

% If you set papersize explicitly, activate the following three lines:
%\special{papersize = 8.5in, 11in}
%\setlength{\pdfpageheight}{11in}
%\setlength{\pdfpagewidth}{8.5in}

% If you use natbib package, activate the following three lines:
% \usepackage{subfigure,graphicx}
% \usepackage{amsmath,amsfonts,amssymb}
% \usepackage{rotating}
% \usepackage[hyperfootnotes=false]{hyperref}
% \usepackage{amsthm}
% \usepackage[ruled,vlined]{algorithm2e}
% \usepackage[table]{xcolor}
% \usepackage{color,colortbl,tabularx}
% % \usepackage{natbib}
% \usepackage[english]{babel}
% \usepackage{booktabs}
% \usepackage{algpseudocode} % For algorithmicx
% \usepackage{float}
% \usepackage{xcolor}
% \usepackage[margin=0.8in]{geometry}
% \usepackage{subcaption}
% \usepackage{multirow}
% \usepackage[round]{natbib}
% \renewcommand{\bibname}{References}
% \renewcommand{\bibsection}{\subsubsection*{\bibname}}

% % % If you use BibTeX in apalike style, activate the following line:
% % %\bibliographystyle{apalike}
\onecolumn
% \begin{document}
% \newpage
\aistatstitle{Supplementary Materials: \\Generating Tabular Data Using Heterogeneous Sequential Feature Forest Flow Matching 
}
\small The code to reproduce the plots and tables in this work can be found at \url{https://github.com/AngeClementAkazan/Sequential-FeatureForestFlow/tree/main}
\begin{figure}[H]
    \centering
    \includegraphics[width=0.95\linewidth]{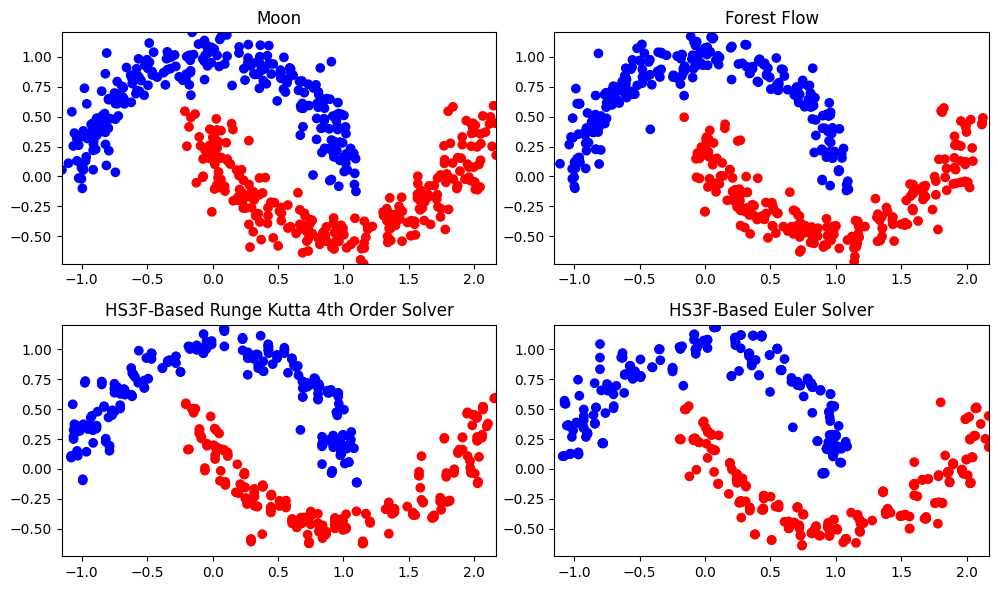}
    \caption{500 samples  2-Moons data set with and its generated version from HS3F-Euler, HS3F-Rg4 and ForestFlow}
    \label{fig:enter-label}
\end{figure}
This part is structured as follows:
\begin{itemize}
\item The section (\ref{SeqFFF}) details the Continuous Sequential Feature Forest Flow Matching and provides clear details about its extension, the Heterogeneous Sequential Feature Forest Flow Matching.
\item The section (\ref{UCI}) provides details about the data sets used in this study.
    \item Section (\ref{barplotss}) contains the bar plots of all metrics per data set for each method.
\end{itemize}

% \usepackage{aistats2025}
% \chapter{Appendix}
\label{appendix}
\section{More Details About the Conditional Flow Matching Concept}\label{flm}
%\subsection{Conditional Flow Matching (CFM) Framework}
\subsection{Flow Matching}
Let us assume that the data space is $\mathbb{R}^n$, with data points $x_1$ following the unknown distribution $q_1$ over $\mathbb{R}^n$,  and that the noisy data point is  $x_0 $ which follows the distribution $p_0$.
The motivation behind flow matching is that there exists a time-dependent probability path $p_t$ such that $p_{t=0}=p_0$ and $p_{t=1}=q_1$, and the goal is to find this probability path.
Flow matching assumes that given $x_0$, $p_t$ is induced by a time-dependent velocity vector field $v_t$, which, over time ($t\in [0,1]$),  directs the trajectory of a time-dependent map $\phi_t$ that pushes, via the push-forward relationship (see Eq.(\ref{flw})), the noisy distribution $p_0$ to $p_t$. 
$\phi_t$ in this case is the flow induced by $v_t$. Given $x_t \sim p_t$, the relationship between the velocity vector field $v_t(x_t)$ and  its induced flow $\phi_t$  can  formally be defined by Eq.(\ref{vct_}) which is an ordinary differential equation (ODE) which specifies that $v_t$ defines how quickly and in which  direction $\phi_t$ evolves over time starting from the noisy data $x_0$: \begin{equation} \dfrac{\partial \phi_t (x)}{dt}= v_t(\phi_t(x)); \quad  \phi_0(x_0)=x_0\sim p_0, \, t\in[0,1].\label{vct_}\end{equation}
The push-forward relationship defined by $\phi_t$  between $p_0$ and $p_t$  is represented as: \begin{equation} 
p_t(x_t)= [\phi_t]_\# (p_0)= p_0(\phi_t^{-1}(x_t)) \cdot det \left| \dfrac{\partial \phi_t^{-1} (x_t)}{\partial x} \right|.
\label{flw}
\end{equation}
When the Eq.(\ref{flw}) and Eq.(\ref{vct_}) are satisfied for a given velocity vector field $v_t$ and its corresponding flow $\phi_t$ which pushes a distribution $p_0$ to the  probability path $p_t$ at time $t\in [0,1]$, then $v_t$ is said to generate  $p_t$. Formally, $v_t$ generates $p_t$ if it satisfies the continuity equation: \begin{equation}
    \frac{{\partial p_t(x)}}{{\partial t}} + \text{div}(p_t(x)v_t(x)) = 0. \quad
\label{conteq}
\end{equation} 
From [Eq.(\ref{vct_}), Eq.(\ref{flw}) and Eq.(\ref{conteq})], we can conclude that the knowledge of a velocity vector field $v_t$ uniquely determines the probability path $p_t$. Therefore, instead of directly determining the right probability path $p_t$ so that $p_1\approx q_1$, which can be very challenging, the flow matching framework intends to approximate the velocity vector field $v_t$ that induces $p_t$. 
Given this information, 
the main objective of flow matching is to regress a neural network  \( v_\theta: [0, 1] \times \mathbb{R}^d \rightarrow \mathbb{R}^d
\) against the velocity vector field $v_t$ by minimizing the loss function $\mathcal{L}_{FM}$ (see Eq.(\ref{flwObj_})), in order to determine the  flow $\hat{\phi}_t$ induced by the trained neural network, which defines a probability path $p_t$ from $p_0$ to $p_1$ which is similar to $q_1$, through the push-forward relationship(Eq.(\ref{flw})).
The training loss  function of flow matching is defined as follows:
 \begin{equation}
    \mathcal{L}_{FM}(\theta) = E_{t, p_t(x)} \|v_t(x) - v_\theta(x)\|^2 
\label{flwObj_}.
\end{equation}
Thereafter, Eq.(\ref{vct_}) has to be  solved numerically (using the trained $v_\theta$)   in order to determine the  flow estimate $\hat{\phi}_t$ that pushes, through the push-forward relationship(Eq.(\ref{flw})), $p_0$ to   $p_1 \approx q_1$. 

Unfortunately, this flow-matching training objective is intractable because we do not have knowledge of suitable $v_t$ and $p_t$. Consequently, we can not have a suitable flow estimate $\hat{\phi}_t$ from this framework.
    % \newpage
\subsection{Conditional Flow Matching}To bypass this aforementioned tractability challenge, 
 \cite{lipman2023flow,tong2023improving} suggested to construct $p_t$ as a mixture of simpler conditional probability path $p_t(x|z)$ which varies with some conditioning feature $z$ that depends on $x_1$, such that $p_0(x|z)=p_0$ and $p_1(x|z)=q_1(x)$. $p_t(x|z)$ is assumed to be  generated by a conditional vector field $v_t(x|z)$ that induces a conditional flow $\phi_t(x|z)$ (from $x_0$ to $x_t$).
 % is some distribution centered around $z$ e.g. $p_1(x|z)=\mathcal{N}(x|z,\sigma I)$.
 \cite{lipman2023flow,tong2023improving}  defined $p_t$ as  the marginalization of  $p_t(x|z)$ over $q(z)$, and determined a suitable velocity vector field $v_t$ based on $v_t(x|z)$ that generates $p_t$ ($v_t$ satisfies the continuity  Eq.(\ref{conteq})), as follows:
\begin{align}
p_t(x)= \mathbb{E}_{q(z)}[p_t(x|z)] &\quad  \text{and} \quad v_t(x)=\mathbb{E}_{q(z)}[v_t(x|z)\dfrac{p_t(x|z)}{p_t(x)}].
\label{pdfpath}
\end{align}
Unfortunately, the expression of $v_t(x)$  Eq.(\ref{pdfpath}) cannot be used directly to solve Eq.(\ref{flwObj_}) because the true $p_t(x)$ remains unknown. Therefore, computing an unbiased estimate of the flow matching loss function $\mathcal{L}_{FM}$ (Eq.(\ref{flwObj_})) is not possible. To overcome this challenge,  \cite{lipman2023flow} suggest to instead minimize the conditional flow matching objective $\mathcal{L}_{CFM}$ (see Eq.(\ref{CondflwObj_})) and demonstrated that $\mathcal{L}_{CFM}$ and  $\mathcal{L}_{FM}$(Eq.(\ref{flwObj_})) have the same gradient:
\begin{equation}
    \mathcal{L}_{CFM}(\theta) = E_{t, q(z), p_t(x|z)} \|v_{\theta}(\phi_t(x|z))-v_t(x|z)\|^2.
\label{CondflwObj_}
\end{equation}

Therefore, regressing a neural network $v^t_\theta$ against $v_t(x|z)$ through $\mathcal{L}_{CFM}$ yields the same optimal solution as regressing $v^t_\theta$ against  $v_t(x)$ through  $\mathcal{L}_{FM}$.

For specificity purposes, it is worth mentioning that the conditional flow matching paradigm is entirely defined by  the nature of the conditional probability path $p_t(x|z)$, the conditional velocity vector $v_t(x|z)$ and its induced condition flow $\phi_t(x|z)$ $\forall t\in [0,1]$ and also the conditioning feature $z$. 
% As $v_t(x|z)= \dfrac{\partial \phi_t(x_0|z)}{\partial t}$
\subsection{Gaussian Conditional Flow Matching }
Gaussian conditional flow matching (GCFM) is a conditional flow matching concept introduced by \cite{lipman2023flow}, which assumes that the initial sample distribution is a standard Gaussian ($x_0 \sim p_0= \mathcal{N}(x_0|0,I)$),  makes Gaussian assumptions on 
the conditional probability path ($p_t(x|z)= \mathcal{N}(x | \mu_t(z), \sigma_t(z)^2 Id_n))$), and assumes a linear conditional flow $(\phi_t(x_0|z)=\mu_t(z)+ x_0\sigma_t(z))$ so that  $\phi_0(x_0|z)=x_0$ and  $\phi_1(x|z)=x_1$ and $z=x_1$. \cite{lipman2023flow} demonstrated that for $\sigma_t(z)>0$ the unique conditional velocity flow $v_t(x|z)$ that induces $\phi_t(x|z)$ and generates $p_t(x|z)$ and has the form: \begin{equation}
   v_t(x|z)= \dfrac{1}{\sigma_t(z)} \dfrac{\partial \sigma_t(z)}{\partial t}(x-\mu_t(z))+\dfrac{\partial \mu_t(z) }{\partial t}, 
   \label{cond_veloc}
\end{equation}
A variant of the GCFM is the Independent Coupling Flow Matching (ICFM) \cite{tong2023improving}  that provides better and faster flow directions. The ICFM framework assumes that  $x_0$  and $x_1$ are sampled independently and $z=x_0,x_1)$, $p_t(x|x_0,x_1)$ is a  Gaussian distribution with expectation $\mu_t(x_0,x_1)=(1-t)x_0+tx_1$ and standard deviation $\sigma_t(x_0,x_1)=\sigma\in [0,1]$, and a corresponding linear flow $x_t=\mu_t(x_0,x_1)+x\sigma_t(x_0,x_1)$ which is induced by  the conditional velocity vector $v_t(x|x_0,x_1)=x_1-x_0$. 

\subsection{Conditional Flow Matching Algorithm and Motivation for\emph{CS3F}} \label{cs3f_motiv}
\begin{algorithm}[H]
\caption{Conditional Flow Matching Training \citep{lipman2023flow}}
\label{FM_Algor1}
\KwIn{  $x_1$:real dataset, $q$: uniform distribution over  $x_1$, $p_0$: noise distribution, initialized $v^{\theta}_t$}
\While{not converged}{
    $t \sim \mathcal{U}([0,1])$ \;
    $x_1 \sim q(x_1)$ \;
    \textcolor{darkgreen}{$x_0 \sim p_0(x_0)$ \texttt{[sample noise]}} \;
    $x_t = \phi_t(x_0 \mid x_1)$ \texttt{[conditional flow]} \;
    Gradient step with $\nabla_{\theta} \| v^{\theta}_t(x_t) - \dot{x}_t \|^2$ \;
}
\KwOut{$\hat{v}^{\theta}_t$}
\end{algorithm}
\begin{algorithm}[H]
\caption{Conditional Flow Matching Sampling}
\label{FM_Algor2}
\KwIn{Trained model $\hat{v}_t^{\theta}$}
\textcolor{darkgreen}{$z_0 \sim p_0(z_0)$ \texttt{[sample noise]}} \;
Numerically solve ODE $\dot{z}_t = \hat{v}^{\theta}_t(z_t)$ \;
\KwOut{$z_{t=1}$}
\end{algorithm}
Consider that we have an initial dataset $x_1=\{x^1,\dots,x^K\}\in \mathbb{R}^{d\times K}$ having $K$ features, $D_K= \{1,\dots,K\}$ is the set of indices of the features of $x_1$ and a standard Gaussian noisy data set $x_0=\{x^k_0,\dots,x^K_0\}\in \mathbb{R}^{d\times K}$.
Forest flow, as any other CFM-based method, determines the flow $\phi_t$ from a standard Gaussian noise $x_0$, to the real data $x_1$ by numerically solving an ODE (see Eq.(\ref{vct_})) but uses the prediction of the  Xgboost regression model $v^t_{\theta}$ (instead of a neural network as for CFM-based methods) trained on the conditional flow $\phi_t(x|(x_0,x_1))=(1-t)x_0+tx_1, \, (t \in [0,1])$ to  approximate the vector field $v_t$. \newline \newline  One condition for this framework to work is that the initial condition of the flow ODE  (see Eq.(\ref{vct_}))  should be equal or from the same distribution with $x_0$  ($z_0 \sim p_0$). What would happen if the initial condition of the ODE flow was slightly modified from standard Gaussian to Gaussian with expectation $\mu=a\in \mathbb{R}$ and variance $\sigma^2=bI_n\in \mathbb{R}$ for instance?
\emph{We experimentally proved that the performance in term of Wassertein distance will drop significantly. This concern is  the main motivation for the deployment of CS3FM.}

\section{Sequential FeatureForestFlow Matching}
\label{SeqFFF}
\subsection{Continuous Sequential FeatureForestFlow Matching}
\label{cs3FMatching}
\emph{CS3F} was deployed to make data generation more robust to changes in initial conditions.
The\emph{CS3F}is a ICFM-based method that handles data feature-wise generation sequentially. Let us assume that we have an initial dataset $x_1=\{x^1,\dots,x^K\}\sim q_1$ that has $K$ features, $D_K= \{1,\dots,K\}$ the set of indices of the features of $x_1$ and the  standard Gaussian noisy data set $x_0=\{x^k_0,\dots,x^K_0\}\in \mathbb{R}^{d\times K}$ whose distribution is denoted as $p_0$.
We assume that each of  the features $x^k$ follow an unknown distribution $q^k_1$, and that each feature $x^k_0$ follows a standard Gaussian distribution denoted by $p^k_0$. Let us denote by $n_s$ the number of time steps $t$ that are used in the flow process and  assume that $t \in T_{level}=\{t_0, \dots,t_{n_s-2},t_{n_s-1}\}\in [0,1].$
% \{\dfrac{1}{n_s},\dots,\dfrac{n_s-1}{n_s},1\}$. 

Using a feature-wise ICFM framework  $\forall k\in D_K$, we result in a per feature conditional variable  $z^k=(x^k_0,x^k)$, Gaussian conditional probability paths, with their corresponding conditional velocity flows and calculated conditional velocity vectors, which are defined $\forall t \in T_{level}$ as follows.
\begin{align*}
p^k_t(x|z^k)&= \mathcal{N} \left (x|z^k, \mu^k_t=(1-t)x^k_0 + tx^k, \sigma^k _t=0 \right ), \, \,\,  x^k_t=\phi^k_t(x|z^k)=  \mu^k_t \,\,  and \, \, v^k _t(x|z^k)= x^k - x^k_0,  
         \label{flow1}
\end{align*} 

\paragraph{Continuous Sequential Feature Flow Matching (CS3F) Training :}
For $x^k \in x_1$, we create $N=K*n_s$ input data by concatenating conditional flow vectors $x^k_t$ with previous features $\{x^1,\dots, x^{k-1}\}$, $\forall t\in T_{levels}$ and use them to, respectively, train $N$ Xgboost regressor models: $f^{\theta^k _t}(x^k_t,x^1,\dots, x^{k-1})$ $\forall k\in D_K, \, \, \forall t \in T_{levels}$, to learn  the conditional vectors field $v^k _t(x|z^k)=x^k - x^k_0$. For $x^k \in x_1$, the conditional feature flow matching training objective is :\begin{align}
\underset{\theta^k _t}{argmin} \mathbb{E}_{t, q(z^k), p^k_t(x|z^k)} \left \| f^{\theta^k _t}[x^k_t|x^1,\dots, x^{k-1} ]-(x^k-x^k_0)\right \|^2, \, \, k\in D_K \, \textbf{and}\, \,  \forall t \in T_{level}
 % \label{cs3f_}
\end{align} 
It is worth noting that when k=1, the Xgboost model takes only $x^1_t$ as its input.
%, and that  the loss function considered was the mean squared error.
\paragraph{Model Generation:}
After training, $\forall k\in D_K$, we obtain $N$ Xgboost models $(\hat{f}^{\theta^k _t})$   that learned $N$ velocity vectors $v^k _t(x|z^k)$. 
% Let us denote by $F^{Xgb}_k=\{ \hat{f}^{\theta^k_t} | \quad \forall  t \in T_{set} \}$, the set of  $n_s$ Xgboost models that learned the per feature velocity vector field $v^k _t(x|z^k)$, $\forall  t \in T_{set}$.
% We solve Equation\ref{vct}  numerically by using Euler implicit and Runge-Kutta 4th order solvers \citep{akinsola2023numerical} to determine the per feature continuous flow (${\phi_t}^{k}=\Tilde{x}^k_t$) 
% for  $ k\in D_K$.

We first start by determining the flow that pushes the $p^k_0$ standard Gaussian distribution to a distribution $p^k_{1}$ which is similar to  $q^k_1$, via the following ODE: \begin{equation}
    \dfrac{\partial \Tilde{x}^1_t}{\partial t}= \hat{f}^{\theta^1 _t}( z^1_{t}), \, \, z^1_{t}\sim  \mathcal{N}(0,I_n),\, \, \Tilde{x}^1_{t=0}= z^1_{t} , \, \,
    \forall t \in T_{level},
    \label{flow_EQ}
\end{equation}
%The resulting  $\Tilde{v}^1_1$ is the approximated  continuous flow $(\hat{\phi})$ from $x^1_{noise}$ to $x^1$, in  other words, $\hat{\phi}(x^1_{noise})=\Tilde{v}^1 \sim  P_{x^1}$. 

Afterward,  $\forall k\in \{2,\dots,K\}$,  using as input the noisy vector $z^k_{t}\sim  \mathcal{N}(0,I_n)$ concatenated with the previous generated features $ (\Tilde{x}^{1}_1,\cdots,\Tilde{x}^{k-1}_1)$, we numerically solve the following ODE: 

\begin{align}
\dfrac{\partial \Tilde{x}^k_t}{\partial t}&= \hat{f}^{\theta^k _t}( z^k_{t}, \Tilde{x}^{1}_1,\cdots,\Tilde{x}^{k-1}_1), \, \, \, \Tilde{x}^k_{t=0}=z^k_{t} , \, \,
    \forall t \in T_{level}.
\label{vct1__}
\end{align} 
The resulting $\Tilde{v}^k$, is a generated sample whose distribution is similar to  $x^k, \, \, \forall k\in \{2,\dots,K\}$. At the end of this process, we obtain a generated data set $\Tilde{X}_1=\{ \Tilde{x}^1_1, \dots, \Tilde{x}^K_1 \} \sim  p_1 \approx q_1$.

\subsection{Heterogeneous Sequential Feature Forest Flow Matching}
\label{HS3F}
Assuming that the data set $x_1$ contains $N_c$ categorical features and let $I_{cat}$ and $I_{cont}$  be the respective set of indices of categorical and continuous features in $x_1$ ($ D_K= I_{cat}\bigcup I_{cont}$).
Let us denote by $v_{cat}=\{x^{k}   \mid k\in I_{cat}\} \in x_1$, the set of all categorical features in $x_1$.  We train a different Xgboost classifier on each input data $X_{k-1}=\{ x^{1},\dots,x^{k-1}\}$, to learn each categorical feature $x^{k} \in v_{cat}$. The training objective function can be expressed as \begin{align}
\underset{\theta^k}{argmin}\, Classification Loss \left (f^{\theta^k}(X_{k-1}),x^{{k}}  \right)
, \,  \,  x^{k}\in v_{cat}
\end{align} 

We obtained afterwards, $N_c$ different trained  Xgboost classifiers $\hat{f}^{\theta^{k}}$ that respectively learned the $N_c$ categorical features $x^k \in v_{cat}$.
For continuous features, we use the \emph{CS3F} training  for $(K-N_c)\times n_s$ models.

\textbf{Continuous Feature  Generation} Each continuous feature is generated using the \emph{CS3F} data generation method.

 \textbf{Discrete Feature  Generation}
% , the output of the Xgboost classifiers $\hat{f}^{\theta^k}$ are vector of predicted normalized  probabilities (whose some equal 1). 
For each $x^{k} \in V_{cat}$, after training, we use the previously generated features $\{\Tilde{x}^{1},\dots, \Tilde{x}^{k-1}\}$ as input to  the  Xgboost classifiers $\hat{f}^{\theta^k}$ whose output are vector of predicted normalized  probabilities (whose summation equal 1). 
\begin{align}\hat{f}^{\theta^k}\left(   \Tilde{x}^{1},\dots, \Tilde{x}^{k-1} \right)=[Prob(class1),\dots, Prob(Maxclass)]_{\Tilde{x}^{k}}.\end{align}  
We then use the predicted class probabilities as input of  multinomial sampling for generating $\Tilde{v}^k$ (with same number of observations as $x^k$) which follows similar distribution with $x^k$.
The end of this process results in a fully generated data set $\Tilde{X}_1=\{ \Tilde{x}^1, \dots, \Tilde{x}^K \} \sim P_1 \approx q_1$.
 \begin{figure}[H]
    \centering   \includegraphics[width=1.1\linewidth]{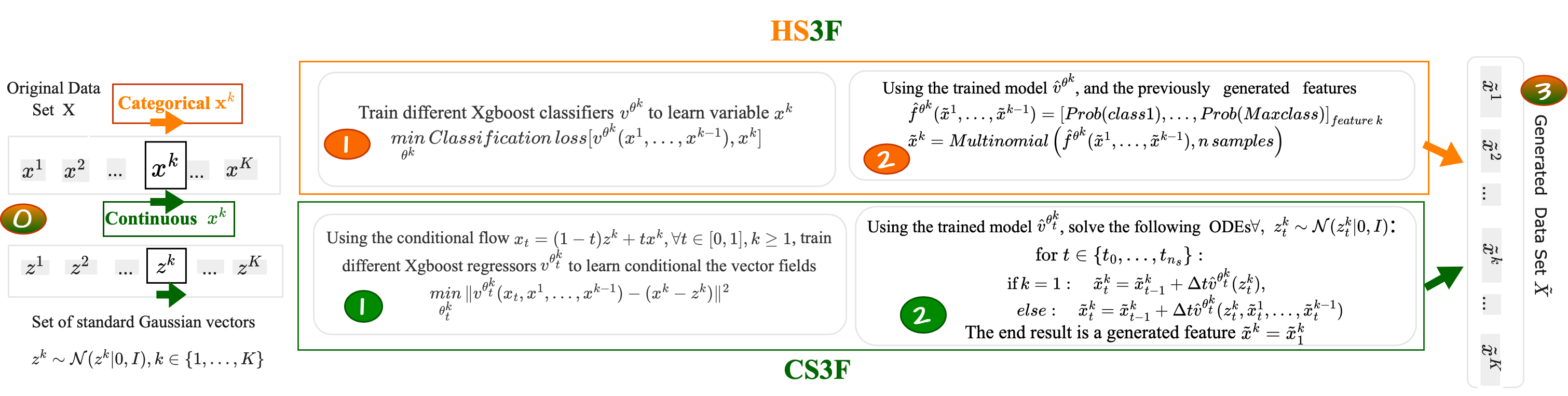}
    \caption{HS3F-based Euler Solver (orange and green rectangle, showing the generation process for both continuous and categorical data) and CS3F-based Euler Solver  (encompasses  the area inside the green rounded rectangle, showing the continuous data generation). $X=\{x^1,\dots,x^K\}$ is the  original  data set and $\{z^1,\dots,z^K\}$ is a set containing  standard Gaussian feature vectors.  The green arrows and steps indicates the generative process continuous features generation  while the orange ones indicates that of categorical features. step 0 and 3 contain both colors, which means that  both methods use these steps except that the categorical feature generation does not use the set of standard Gaussian noise.} 
    \label{cs3f_fig}
\end{figure} 
% \newpage
\section{Data sets}
\label{UCI}
 The  data sets Iris and Wine are from scikit-learn and licensed with the BSD 3-Clause License and the remaining are  imported from UCI  and licensed under the Creative Commons Attribution 4.0
International license (CC BY 4.0). All data sets are openly shared with no restrictions on usage.
\begin{table}[H]
\centering
\caption{Dataset Information}
\begin{tabular}{|l|l|l|c|c|c|}
\hline
\textbf{Order} &\textbf{Dataset} & \textbf{Reference} & \textbf{Row} & \textbf{Column} & \textbf{Output type} \\
\hline
1&Iris & \citep{fisher1988iris} & 150 & 4 & Categorical \\
2&Wine & \citep{aeberhard1991wine} & 178 & 13 & Categorical\\
3&Parkinsons & \citep{little2008parkinsons} & 195 & 23 & Categorical \\
4&Climate Model Crashes & \citep{lucas2013climate_model_crashes} & 540 & 18 & Categorical \\
5&Concrete Compression & \citep{yeh2007concrete_compression} & 1030 & 7 & Continuous \\
6&Yacht Hydrodynamics & \citep{gerritsma2013yacht} & 308 & 6 & Continuous \\
7&Airfoil Self Noise & \citep{brooks2014airfoil} & 1503 & 5 & Continuous \\
8&Connectionist Bench Sonar & \citep{sejnowski_sonar} & 208 & 60 & Categorical \\
9&Ionosphere & \citep{sigillito1989ionosphere} & 351 & 34 & Categorical \\
10&QSAR Biodegradation & \citep{mansouri2013qsar_biodegradation} & 1055 & 41 & Categorical \\
% Bean & \citep{koklu2020bean} & 13611 & 16 & Categorical \\
11&Seeds & \citep{charytanowicz2012seeds} & 210 & 7 & Categorical \\
12&Glass & \citep{german1987glass} & 214 & 9 & Categorical \\
13&Ecoli & \citep{nakai1996a} & 336 & 7 & Categorical \\
14&Yeast & \citep{nakai1996a} & 1484 & 8 & Categorical \\
15&Libras & \citep{dias2009libras} & 360 & 90 & Categorical  \\
16&Planning Relax & \citep{bhatt2012planning_relax} & 182 & 12 & Categorical \\
17&Blood Transfusion & \citep{yeh2008blood_transfusion} & 748 & 4 & Categorical \\
18&Breast Cancer Diagnostic & \citep{wolberg1995breast_cancer} & 569 & 30 & Categorical \\
19&Connectionist Bench Vowel & \citep{deterding_vowel} & 990 & 10 & Categorical \\
% California Housing & \citep{pace1997california_housing} & 20640 & 8 & Continuous \\
20&Concrete Slump & \citep{yeh2009concrete_slump} & 103 & 7 & Continuous \\
21&Wine Quality Red & \citep{cortez2009wine_quality} & 1599 & 10 & Continuous \\
22&Wine Quality White & \citep{cortez2009wine_quality} & 4898 & 11 & Continuous \\
23&Tic-Tac-Toe & \citep{aha1991tic_tac_toe} & 958 & 9 & Categorical \\
24&Congressional Voting & \citep{congressional_voting} & 435 & 16 & Categorical \\
25&Car Evaluation & \citep{bohanec1997car} & 1728 & 6 & Categorical \\
\hline
\end{tabular}
\label{tab:datasets}
\end{table}

\section{Individual Performance Bar Plots Per Data Set}  \label{barplotss}
In this section, we provide the bar plots across all the performance metrics of the methods we used. 
\subsection{Wasserstein Score}
\begin{figure}[H]
    \centering
    \includegraphics[width=1.0\linewidth]{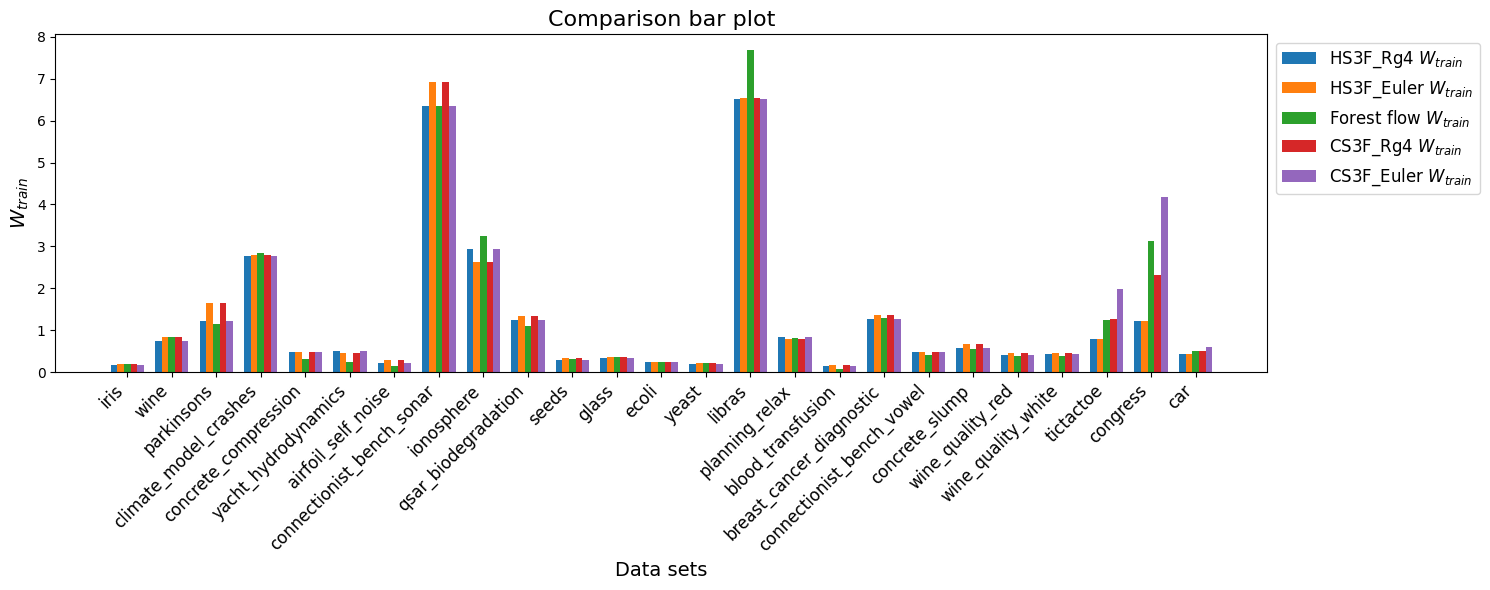}
    \caption{Wasserstein train comparison across datasets}
    \label{fig001}
\end{figure}

\begin{figure}[H]
    \centering
    \includegraphics[width=1.0\linewidth]{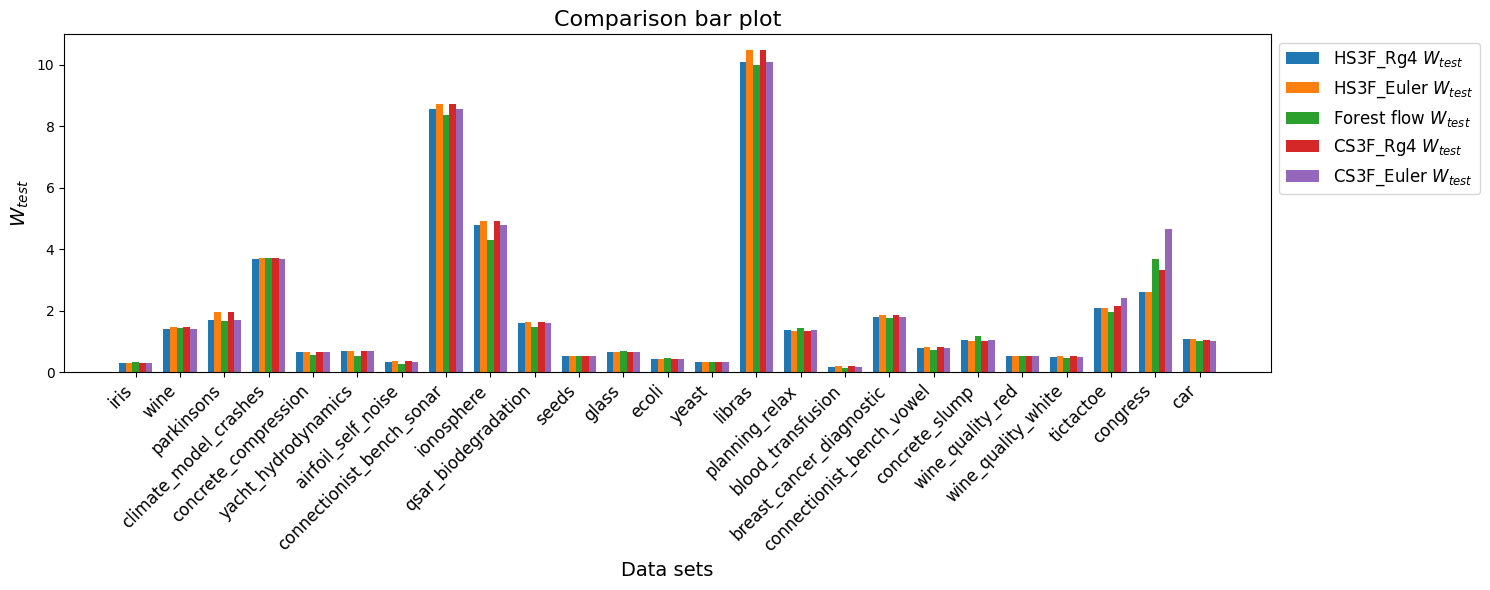}
    \caption{Wasserstein test comparison across generated  datasets}
    \label{fig002}
\end{figure}
\subsection{F1 Score}
\begin{figure}[H]
    \centering
    \includegraphics[width=1.0\linewidth]{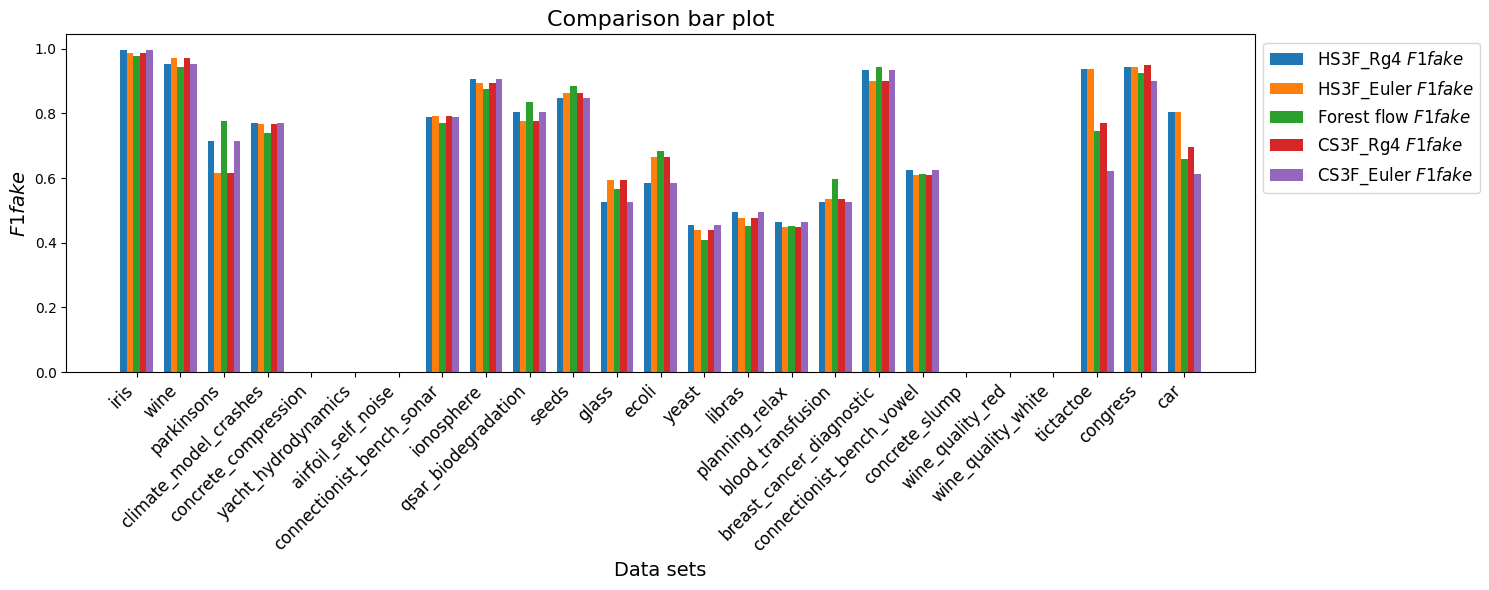}
    \caption{F1 score  comparison of models trained on generated datasets}
    \label{fig1}
\end{figure}

\begin{figure}[H]
    \centering
    \includegraphics[width=1.0\linewidth]{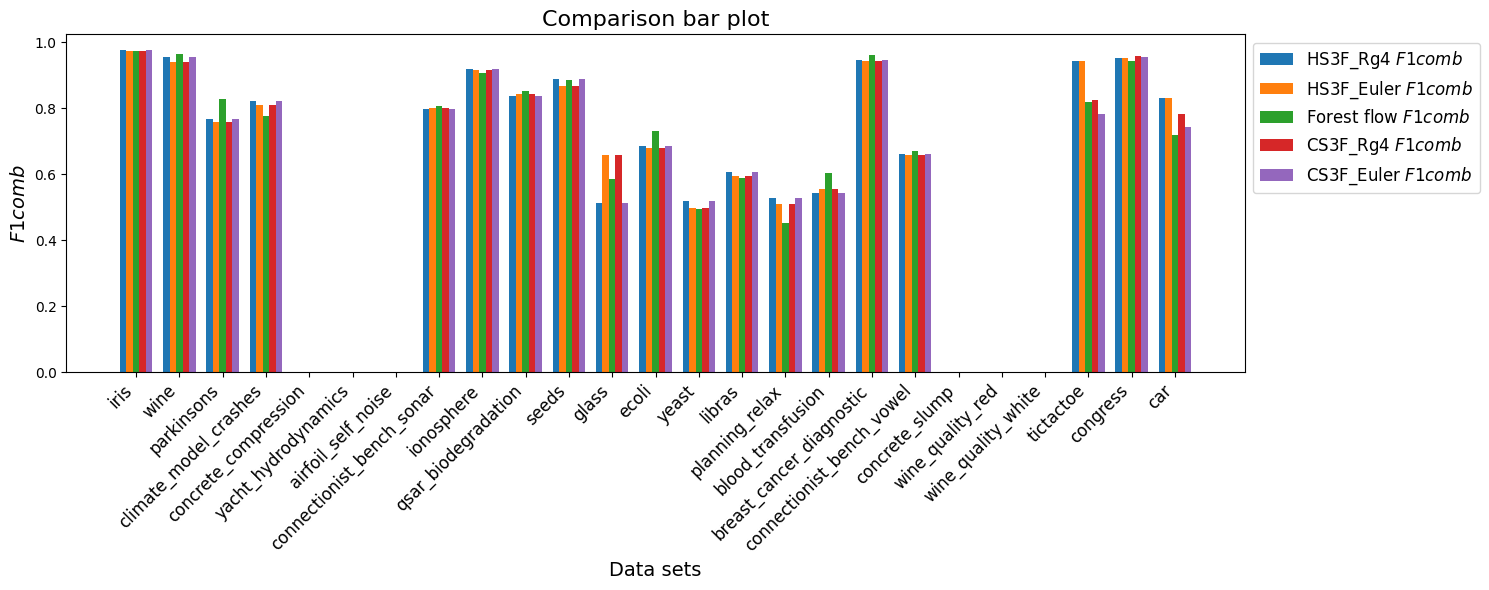}
    \caption{F1 score  comparison of models trained on combined datasets (fake+real)}
    \label{fig2}
\end{figure}

\subsection{Coefficient of determination (R2 Squared)}

\begin{figure}[H]
    \centering
    \includegraphics[width=1.0\linewidth]{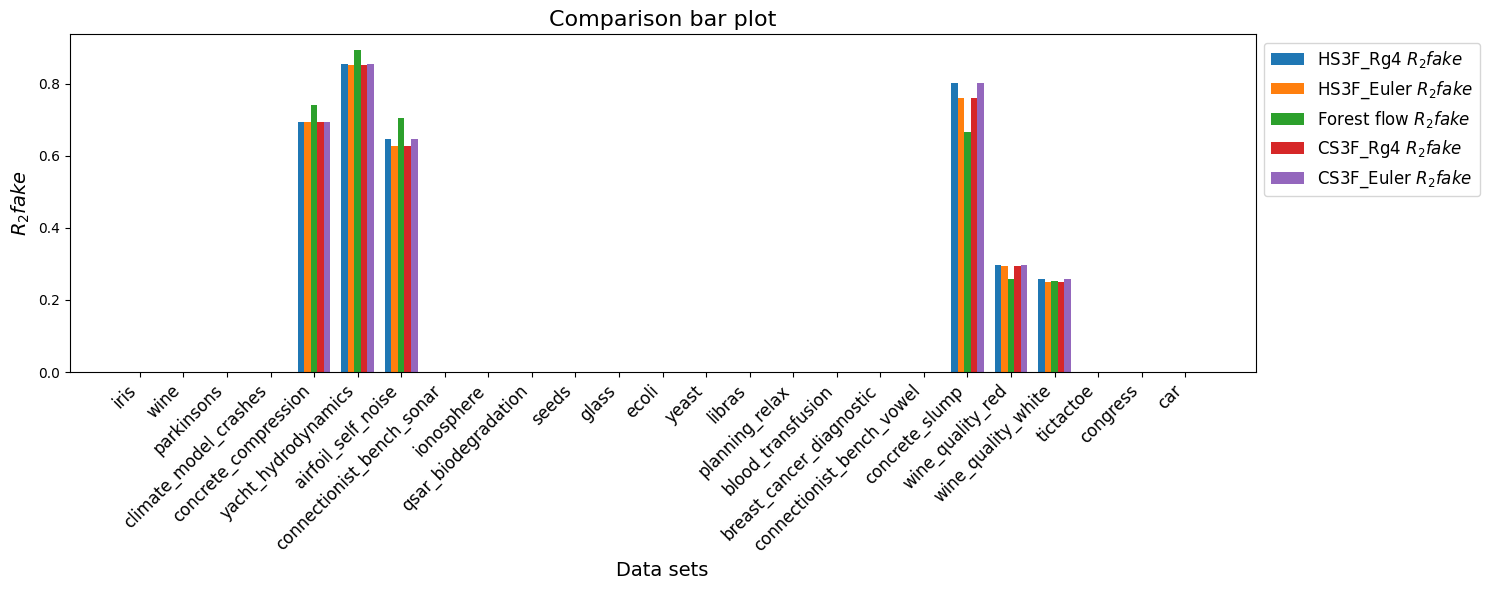}
    \caption{R2  comparison of models trained on generated datasets}
    \label{fig3}
\end{figure}

\begin{figure}[H]
    \centering
    \includegraphics[width=1.0\linewidth]{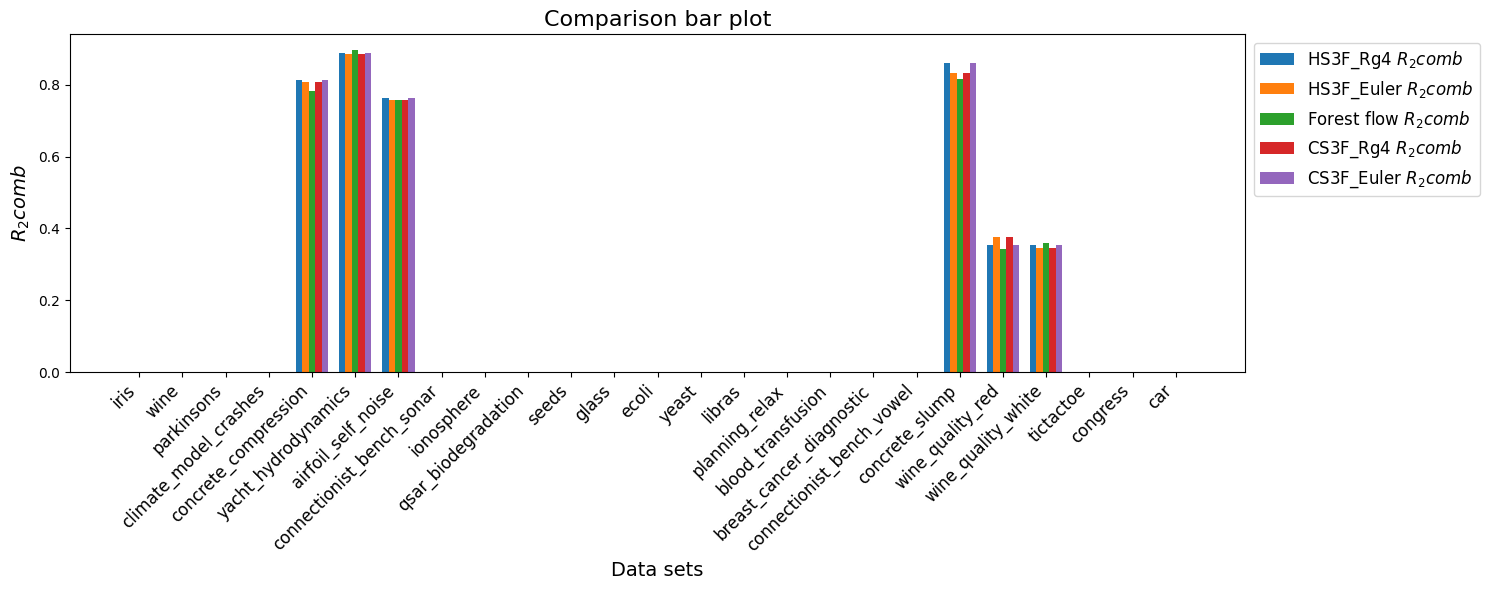}
    \caption{R2  comparison of models trained on combined datasets (fake+real)}
    \label{fig4}
\end{figure}

% \newpage
\subsection{Coverage}

\begin{figure}[H]
    \centering
    \includegraphics[width=1.0\linewidth]{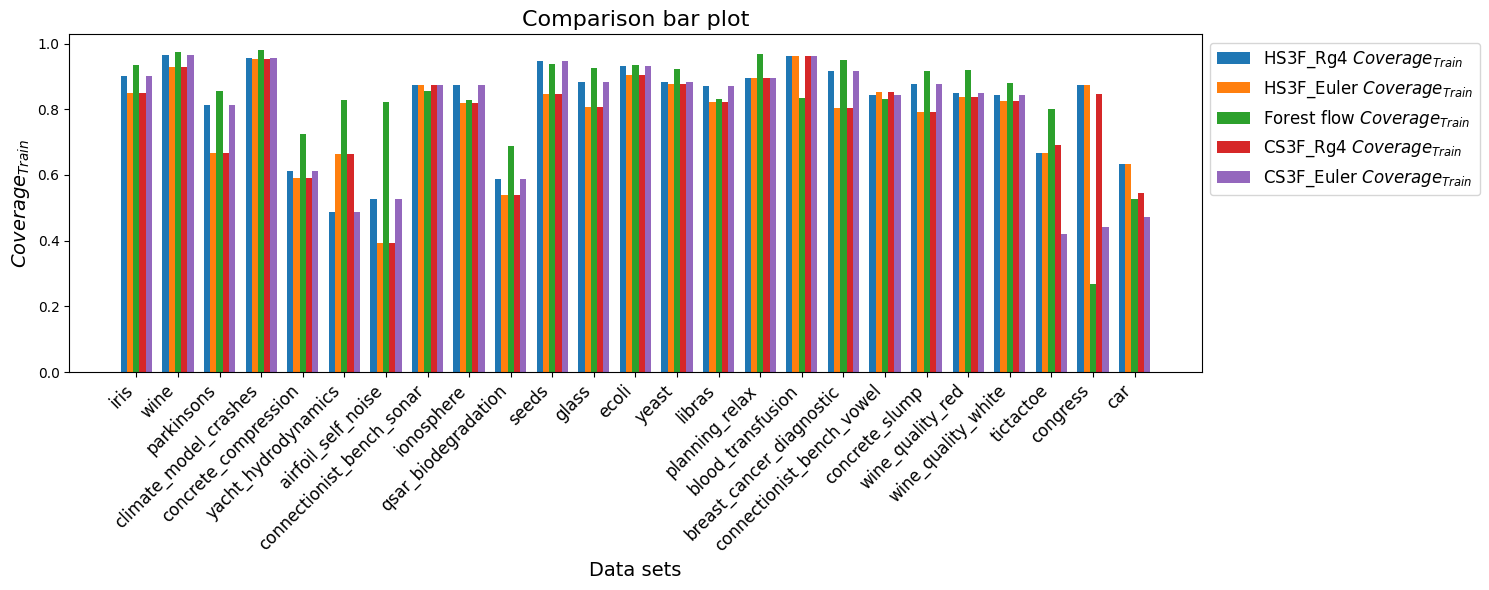}
    \caption{Coverage train comparison across generated  datasets}
    \label{fig5}
\end{figure}

\begin{figure}[H]
    \centering
    \includegraphics[width=1.0\linewidth]{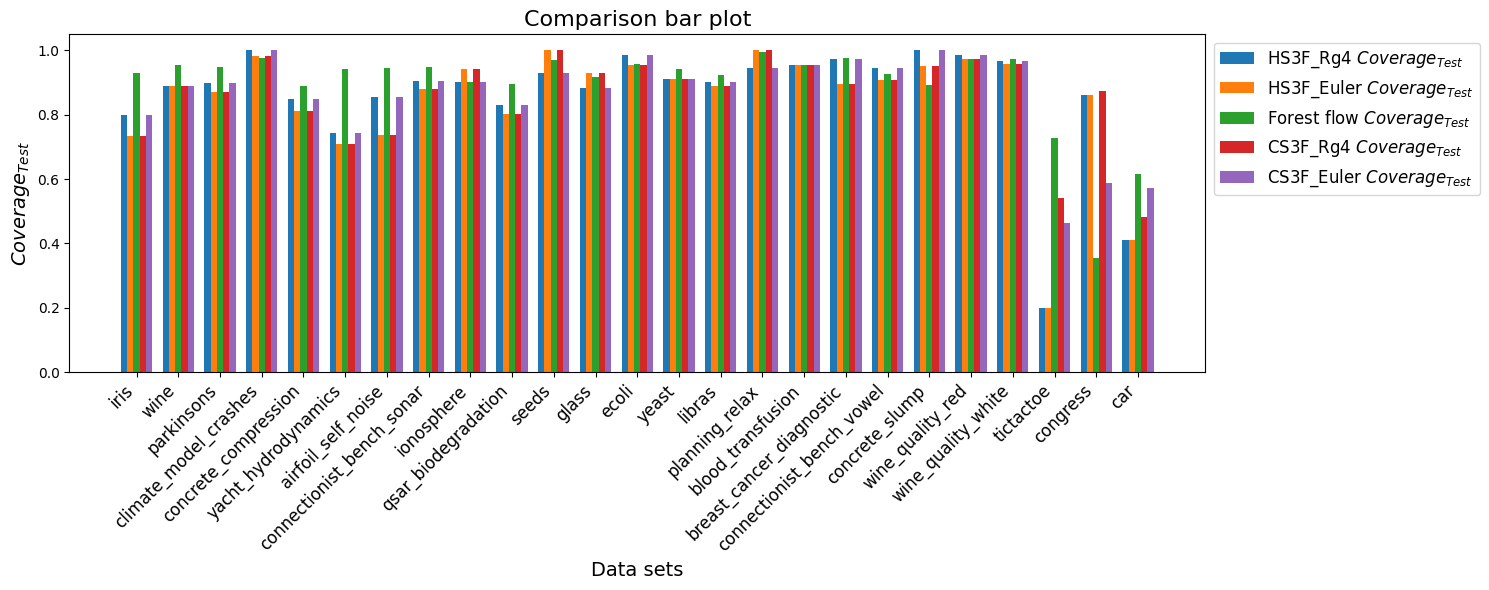}
    \caption{Coverage test comparison across generated  datasets}
    \label{fig6}
\end{figure}
\subsection{Data Generation Running Time}
\begin{figure}[H]
    \centering
    \includegraphics[width=1.1\linewidth]{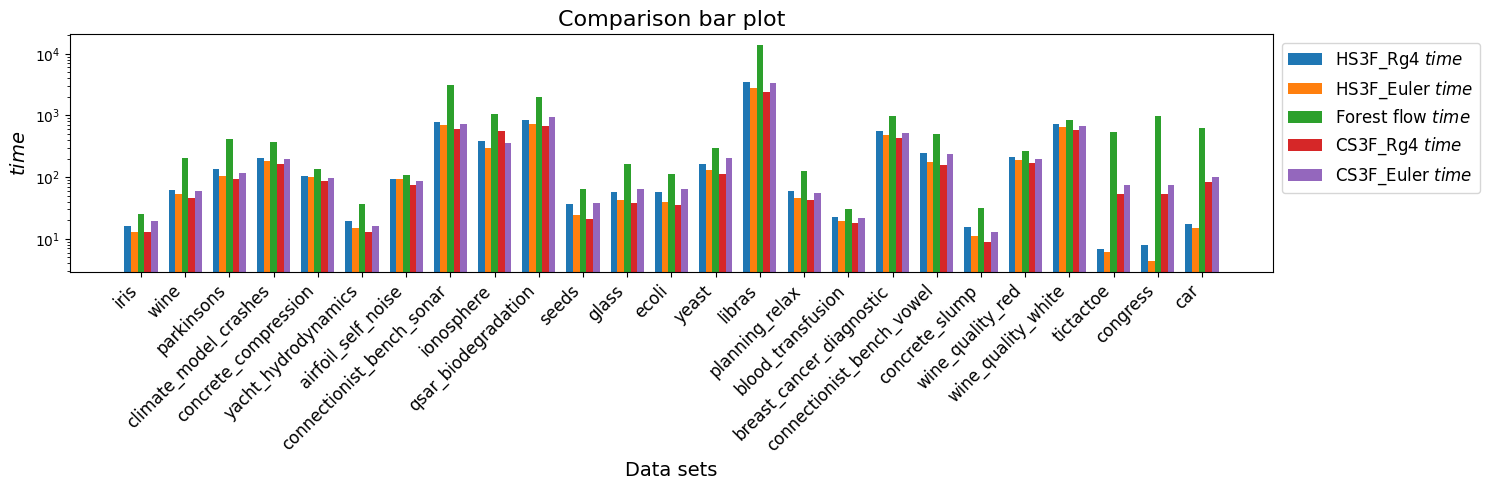}
    \caption{Data generation time comparison per models  across generated  datasets}
    \label{fig6_}
\end{figure}

% \bibliographystyle{abbrvnat}
% \bibliography{bibliography_CIBB_file.bib} 
% \normalsize
% \end{document}

\end{document}